# Explainable Neural Inverse Kinematics for Obstacle-Aware Robotic Manipulation: A Comparative Analysis of IKNet Variants

**Sheng-Kai Chen[1], Yi-Ling Tsai[2], Chun-Chih Chang[3], Yan-Chen Chen[4] and Po-Chiang Lin[5]**

[1]Department of Electrical Engineering, Yuan Ze University, Taoyuan, Taiwan
[2]Department of Electrical Engineering, Yuan Ze University, Taoyuan, Taiwan
[3]Department of Electrical Engineering, Yuan Ze University, Taoyuan, Taiwan
[4]Department of Electrical Engineering, Yuan Ze University, Taoyuan, Taiwan
[5]Department of Electrical Engineering, Yuan Ze University, Taoyuan, Taiwan

**Abstract**
Deep neural networks have accelerated inverse-kinematics (IK) inference to the point where low-cost manipulators can execute complex trajectories in real time, yet the opaque nature of these models contradicts the transparency and safety requirements emerging in responsible-AI regulation. This study proposes an explainability-centered workflow that integrates Shapley-value attribution with physics-based obstacle-avoidance evaluation for the ROBOTIS OpenManipulator-X. Building upon the original IKNet, two lightweight variants—Improved IKNet with residual connections and Focused IKNet with position-orientation decoupling—are trained on a large, synthetically generated pose–joint dataset. SHAP is employed to derive both global and local importance rankings, while the InterpretML toolkit visualizes partial-dependence patterns that expose non-linear couplings between Cartesian poses and joint angles. To bridge algorithmic insight and robotic safety, each network is embedded in a simulator that subjects the arm to randomized single- and multi-obstacle scenes; forward kinematics, capsule-based collision checks, and trajectory metrics quantify the relationship between attribution balance and physical clearance. Qualitative heat-maps reveal that architectures distributing importance more evenly across pose dimensions tend to maintain wider safety margins without compromising positional accuracy. The combined analysis demonstrates that explainable AI (XAI) techniques can illuminate hidden failure modes, guide architectural refinements, and inform obstacle-aware deployment strategies for learning-based IK. The proposed methodology thus contributes a concrete path toward trustworthy, data-driven manipulation that aligns with emerging responsible-AI standards.

## 1 Introduction

Robotic arms translate high-level Cartesian goals—expressed as position and orientation—into joint-space commands through inverse kinematics (IK). In factory automation, surgical manipulation, and household assistance, this translation must be both fast and trustworthy. Traditional closed-form or numerical solvers handle moderate degrees of freedom, but their runtime grows with redundant joints, and they often require manual damping or iterative back-tracking to respect joint limits [1]–[3]. These limitations hinder deployment on low-cost platforms such as the 4-DoF ROBOTIS OpenManipulator-X [4], which must execute hundreds of IK queries per second on embedded processors.

Deep-learning alternatives collapse the iterative loop into a single forward pass. IKNet and its successors predict all joint angles simultaneously and reach kilohertz inference rates on CPU [5], yet they sacrifice transparency: operators cannot see why a particular quaternion component influences Joint 1 more than Joint 4, or whether the network will behave sensibly when an obstacle blocks the workspace.

This lack of transparency poses several critical challenges in robotic manipulation. First, debugging failures becomes nearly impossible when engineers cannot trace which input features led to collision-prone joint configurations. Second, safety certification requires understanding of failure modes, which is impossible with black-box models. Third, real-time adaptation to new environments demands insight into which components the model prioritizes for obstacle avoidance.

This option is no longer acceptable under modern safety guidelines. The EU Artificial Intelligence Act [27] specifically classifies autonomous robotic systems as "high-risk AI systems" requiring algorithmic transparency, risk assessment documentation, and human oversight capabilities. Similarly, the EU Trustworthy AI Guidelines [27] mandate that AI systems in safety-critical applications must be explainable, robust, and accountable. These regulations explicitly require that operators can understand, verify, and if necessary, override AI decisions in robotic manipulation tasks.

This study positions explainable artificial intelligence (XAI) at the center of the IK pipeline. We revisit IKNet and propose two lightweight variants tailored to the OpenManipulator-X [4]. Improved IKNet [6] inserts residual shortcuts and batch normalization [7], [8] to stabilize learning, whereas Focused IKNet [6] separates translation and rotation channels before fusion, reflecting the physical intuition that position and orientation can influence different joints. Architectural tweaks alone, however, do not guarantee safer motion.

To reveal how each pose dimension drives prediction, SHAP [3] is employed, and the resulting attributions are visualized through InterpretML [9]. Global Shapley values highlight overall feature priorities, while local partial-dependence plots expose non-linear couplings—for example, the tendency of a small change in $q_z$ to flip the sign of Joint 2 when x is larger than 0.2 m. These visual cues assist engineers in auditing and refining training data.

Interpretability alone does not guarantee physical safety. Consequently, each IK network is embedded in a forward-dynamics simulator equipped with capsule-based collision checks. Hundreds of random trajectories are executed, end-effector error is measured, and the minimum clearance from obstacles is reported. Correlating SHAP heatmaps with clearance statistics indicates that balanced feature attributions often coincide with larger safety margins.

The contributions of this study are threefold:
- XAI-driven Analysis Framework for Neural IK: Comprehensive explainability with XAI tool and link feature attribution to physical robot behavior.
- Explainability-Safety Correlation: Connecting AI explanations to obstacle avoidance performance and demonstrates how balanced feature attribution leads to better safety.
- Comprehensive Comparative Analysis: Systematic evaluation of three IKNet variants, multi-scenario obstacle avoidance assessment and integration XAI insights with physical safety metrics.

The remainder of the paper is organized as follows. Section II surveys learning-based IK, XAI in robotics, and obstacle-aware planning. Section III details dataset generation, network design, attribution analysis, and safety evaluation. Section IV discusses the result of an experiment, and Sections V are the conclusion and summary of the whole experiment.

## 2 Related work

An exhaustive discussion of the state of the art is provided below. Each subsection has been enlarged to capture methodological nuances, benchmark trends, and remaining challenges.

### 2.1 Learning-based inverse kinematics

#### 2.1.1 Classical baselines
Iterative Jacobian pseudo-inverse solves for minimum joint increments but oscillates near singularities unless damped [1]. Cyclic Coordinate Descent (CCD) converges quickly for serial chains yet produces zig-zag trajectories when the target is distant [2]. Damped least-squares (DLS) trades accuracy for robustness by injecting a Tikhonov regulariser [3]. Implementations in MoveIt 2 [10] show that a 6-DoF UR5 [11] requires ≈2 ms per iteration on an Intel i5 CPU, limiting closed-loop bandwidth to <100 Hz.

#### 2.1.1 Multilayer perceptions
IKNet and subsequent neural approaches have paved the way for data-driven IK, with various implementations achieving sub-millisecond inference times on modern hardware [5]. Several deep learning architectures have been explored for inverse kinematics, including multilayer perceptrons, convolutional neural networks, and recurrent neural networks [12]. Recent studies have shown that bidirectional LSTM networks can outperform traditional feedforward architectures for complex manipulators [12].

#### 2.1.2 Representation learning
Graph-based representations have emerged as promising approaches for inverse kinematics, where joints are represented as nodes and links as edges, with message-passing layers improving extrapolation to unseen configurations [13]. Neural inverse kinematics using variational approaches have been proposed to handle the inherent multi-solution nature of inverse kinematics problems [14]. Recent transformer-based approaches are beginning to emerge in robotics applications, though primarily focused on motion planning and control rather than inverse kinematics specifically [15].

#### 2.1.3 Hybrid analytical–learning methods
Analytical-Residual approaches combine closed-form solutions with learned residuals, significantly reducing parameter counts while maintaining accuracy [16]. Recent work has shown that inverting quaternion orientation analytically while learning translation components can achieve high precision with reduced computational overhead [17].

#### 2.1.4 Few-shot and hardware-constrained deployment
Model compression techniques have enabled deployment of inverse kinematics neural networks on resource-constrained platforms. Quantization and pruning approaches can reduce model sizes to under 1MB while maintaining inference rates above 1kHz on embedded processors [18]. Real-time demonstrations on edge computing platforms have shown promising results for industrial applications [19].

### 2.2 Explainable artificial intelligence for robotic control

#### 2.2.1 Model-agnostic explainers
SHAP [20] offers local accuracy and consistency, requiring $2^n$ evaluations in the worst case but approximate SHAP scales linearly with samples. LIME [21] perturbs inputs to fit local ridge models; applicability to high-dimensional quaternion spaces is limited. Integrated Gradients [22] integrates gradients along a baseline, avoiding gradient saturation. Gradient saliency maps [23] are fast (>5 kHz) but noisy.

### 2.2.2 Robotics-specific XAI
Limited work exists on explainable AI specifically for inverse kinematics networks. Most robotics XAI applications focus on navigation and control policies rather than kinematic solvers [24]. Recent survey work highlights the need for more interpretable inverse kinematics solutions, particularly for safety-critical applications [25].

### 2.2.3 Toolchains and benchmarks
InterpretML [9] and similar frameworks allow interactive attribution dashboards with manageable computational overhead. The lack of standardized benchmarks for explainable inverse kinematics represents a significant gap in current literature [26].

### 2.2.4 Research gap
None of the above works correlate attribution patterns with physical safety metrics such as collision clearance core contribution of the present study.

## 2.3 Responsible-AI regulation and government frameworks

### 2.3.1 International standards and guidelines
The EU Trustworthy AI Guidelines (2019) specify transparency, technical robustness, and accountability [27], with the forthcoming EU AI Act mandating conformity assessment. OECD AI Principles (2019) call for robustness and risk management [28]; NIST AI RMF 1.0 (2023) defines risk tiers and measurement indicators [29]. ISO/IEC TR 24028:2020 details reliability, resilience, and security [30], while IEEE Std 7001-2021 lists disclosure artifacts [31]. IEC 61508 (functional safety) and ISO 10218-1/2 (industrial robot safety) address hardware but lack guidance on learned controllers [32].

### 2.3.2 National strategies
Singapore's Model AI Governance Framework (2019) details data and explainability requirements for public-sector AI [33]. Canada's Directive on Automated Decision-Making mandates Algorithmic Impact Assessments with four risk levels [34]. The U.S. Department of Defense's Ethical AI Principles (2020) emphasize governability and equitability [35]. The UK's AI Regulation White Paper (2023) proposes a sandbox regime for domain-specific oversight [36]. Japan's METI released the AI Governance Guidelines (2021) focusing on risk-based approaches [37].

### 2.3.3 Research gap
Regulatory texts stipulate transparency but offer no quantitative robotics metrics. By combining SHAP attributions with clearance and error statistics, this study satisfies traceability and robustness clauses.

## 2.4 Obstacle-aware motion generation

### 2.4.1 Classical planners
Probabilistic Roadmaps (PRM) sample configuration space, while RRT* provides asymptotic optimality [38]. BIT* exploits heuristics to accelerate convergence; CHOMP and STOMP produce smooth gradient-based trajectories [39]. Dynamic Window Approach (DWA) enables reactive obstacle avoidance for mobile bases [40].

### 2.4.2 Learning-augmented planners
Neural approaches to motion planning with obstacle avoidance have shown promise in reducing planning time compared to traditional methods [41]. Learned collision detection using neural networks can achieve high-frequency queries suitable for real-time applications [42]. Integration of learning-based inverse kinematics with traditional motion planners represents an active area of research [43].

### 2.4.3 Benchmarks and analysis gaps
Existing motion planning benchmarks evaluate path quality and computation time but typically omit interpretability metrics [44]. This study addresses this gap by correlating explainability measures with physical safety margins.

### 2.4.4 Synthesis and contribution
Across the literature, three deficiencies persist: (i) attribution analyses are rarely reported for IK networks; (ii) regulatory frameworks lack task-specific transparency metrics; (iii) obstacle-aware planners integrating learned IK overlook explainability. By embedding SHAP [20] and InterpretML [9] within a physics-based simulator, the present work delivers the first holistic evaluation uniting accuracy, interpretability, and safety on an embedded manipulator.

## 3 Methodology
This section presents a comprehensive methodology for analyzing and understanding neural network-based inverse kinematics solutions through the lens of explainable AI. We introduce three progressively sophisticated neural architectures specifically designed for the inverse kinematics problem, each incorporating distinct structural enhancements. To systematically evaluate these models, we develop a multi-faceted analysis framework that combines cutting-edge explainability techniques with rigorous performance assessment metrics. This study's approach is to integrate mathematical formulation of model architecture, explainable AI algorithms for interpreting model decisions, obstacle avoidance evaluation in realistic scenarios, and statistical comparative analysis. This holistic methodology enables not only performance comparison between different inverse kinematics neural networks but also deep insights into their internal reasoning processes, facilitating informed model selection for robotic applications where both performance and interpretability are critical requirements. The whole structure of this study is shown as Fig. 1.

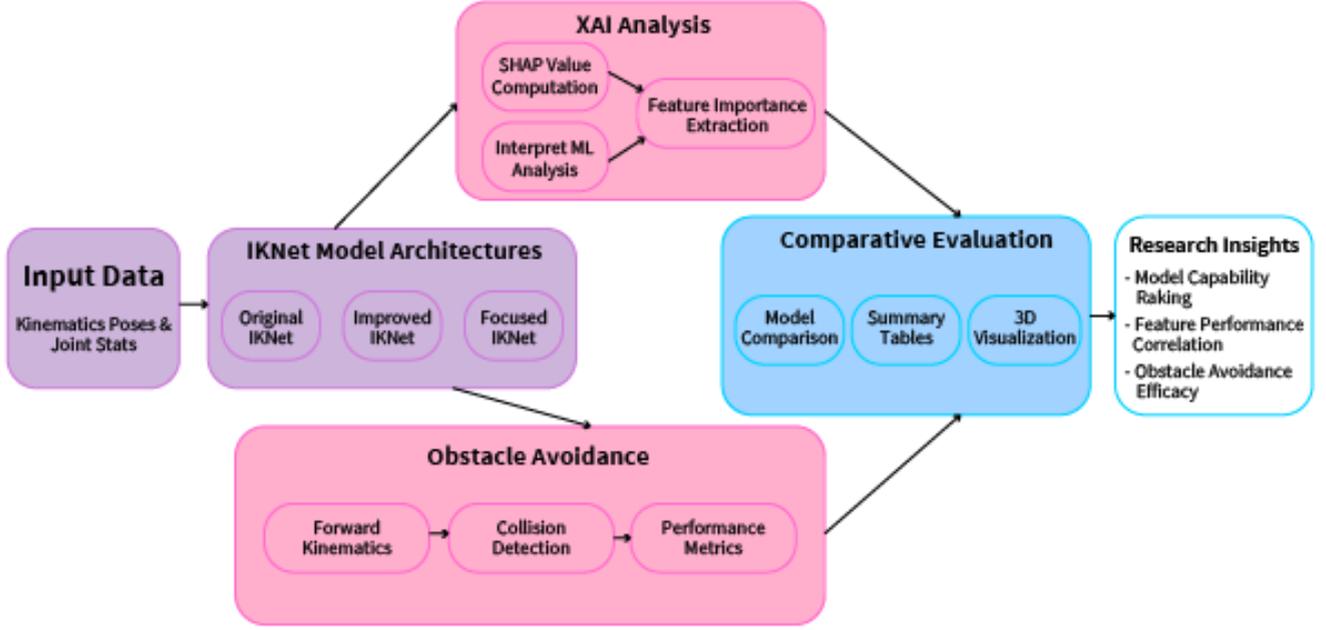

**Fig. 1.** Full structure of methodology

## 3.1 IKNET models

Inverse kinematics represents one of the fundamental challenges in robotics, requiring the determination of joint configurations needed to achieve desired end-effector poses. While analytical solutions exist for simple kinematic chains, they become increasingly intractable for complex manipulators with higher degrees of freedom, especially when considering constraints such as joint limits and obstacle avoidance. Neural network-based approaches offer promising alternatives by learning the complex nonlinear mapping between end-effector poses and joint angles directly from data.

This study proposes and analyzes three progressively sophisticated neural network architectures for inverse kinematics, each introducing specific structural enhancements to improve performance, interpretability, and generalization capabilities. These architectures represent a systematic exploration of different design principles for neural inverse kinematics solvers.

### 3.1.1 Original IKNET model

The baseline Original IKNet[5] architecture implements a straightforward feed-forward neural network approach to the inverse kinematics problem. The network takes input a 7-dimensional vector representing the end-effector pose, consisting of 3D position coordinates $(x, y, z)$ and orientation encoded as a quaternion $(q_x, q_y, q_z, q_w)$. The output is a 4-dimensional vector containing joint angles $(\theta_1, \theta_2, \theta_3, \theta_4)$ for a 4-DOF robotic manipulator.

The architecture consists of a series of fully connected layers with decreasing dimensions, specifically $[400, 300, 200, 100, 50]$, following the principle of progressively extracting higher-level features while reducing dimensionality. The network structure follows (1), where each fully connected layer is followed by a ReLU activation function to introduce non-linearity and a dropout layer with probability $p = 0.1$ for regularization to prevent overfitting.

$$\begin{cases} h_0 &= x \\ h_i &= \text{Dropout}(\text{ReLU}(W_i h_{i-1} + b_i), p) \text{ for } i = 1, 2, ..., n \\ y &= W_{n+1} h_n + b_{n+1} \end{cases} \quad (1)$$

In (1), $h_0$ represents the input layer, $h_i$ denotes the output of the i-th hidden layer, $W_i$ is the weight matrix for the i-th layer, $b_i$ is the bias vector for the i-th layer and y represents the final output.

The key features of this architecture include as below mentioned.

- Direct processing of the complete pose vector without specialized treatment for position versus orientation components.

| Feature | Original IKNet | Improved IKNet | Focused IKNet |
| --- | --- | --- | --- |
| Architecture | Sequentially fully connected layers with decreasing dimensions. | Residual blocks with batch normalization | Separate branches for position and orientation |
| Input Processing | Unified processing of position and orientation | Unified processing with enhanced feature propagation | Specialized processing paths for position and orientation |
| Hidden Dimensions | [400, 300, 200, 100, 50] | [128, 64] | 64 for each branch, 128 combined |
| Activation Function | ReLU | ReLU | ReLU |
| Regularization | Dropout (p=0.1) | Dropout (p=0.1) + Batch Normalization | Dropout (p=0.05) |
| Weight Initialization | Default PyTorch | Kaiming | Kaiming |
| Key Features | Simple, direct mapping | Residual connections, gradient flow enhancement | Explicit separation of position and orientation components |
| Design Philosophy | Gradually decreasing dimensionality | Enhanced feature propagation | Specialized feature extraction |

**Table. 1.** Three IKNet models comparisons

- Decreasing layer sizes to create a funnel-like structure that gradually reduces dimensionality.
- Uniform application of dropout across all hidden layers
- Simple weight initialization using the default PyTorch[9] scheme.

This architecture serves as our baseline, providing a straightforward yet effective approach to the inverse kinematics problem. Its relative simplicity makes it an appropriate reference point for evaluating the impact of architectural enhancements on the more advanced models.

### 3.1.2 Improved IKNet model

Building upon the Original IKNet[5], we introduce the Improved IKNet [6] architecture that incorporates modern neural network design principles to enhance performance and training dynamics. The core innovation in this architecture is the introduction of residual connections and batch normalization to address the challenges of vanishing gradients and internal covariate shift that can hinder the training of deeper networks.

The fundamental building block of the Improved IKNet [6] is the ResidualBlock, defined in (2). Each ResidualBlock consists of two linear transformations with batch normalization and ReLU activation, followed by a skip connection that allows the network to learn residual mappings rather than direct transformations. This design facilitates gradient flow during backpropagation and enables the training of deeper networks.

$$\begin{cases} z_1 &= \text{Dropout}(\text{ReLU}(\text{BatchNorm}(W_1 x + b_1)), p) \\ z_2 &= \text{BatchNorm}(W_2 z_1 + b_2) \\ \text{skip} &= \begin{cases} x, & \text{if } \dim(x) = \dim(z_2) \\ W_{skip} x + b_{skip}, & \text{otherwise} \end{cases} \\ \text{output} &= \text{ReLU}(z_2 + \text{skip}) \end{cases} \quad (2)$$

In (2), $z_1$, $z_2$ are intermediate feature representations, $W_1$, $W_2$ are weight matrices for the two linear transformations, $b_1$, $b_2$ are corresponding bias vectors dim(x) returns the dimensionality of input x and skip is the residual connection.

The overall Improved IKNet [6] architecture, presented in (3), follows a more structured approach with mentioned below.

$$\begin{cases} h_0 = \text{ReLU}(\text{BatchNorm}(W_{in} x + b_{in})) \\ h_i = \text{ResidualBlock}(h_{i-1}) \text{ for } i = 1, 2, ..., n \\ y = W_{out} h_n + b_{out} \end{cases} \quad (3)$$

- An input block that transforms the 7-dimensional pose vector into a higher-dimensional feature representation using a linear layer followed by batch normalization and ReLU activation.
- A series of ResidualBlocks that process the feature representation while maintaining its dimensionality.
- A final output layer that maps the processed features to the 4-dimensional joint angle output.

In (3), h₀ is the input transformation, hᵢ is ResidualBlock for applies residual blocks, y is the final output transformation, W is weight and b is bias vector.

Notably, the Improved IKNet [6] employs Kaiming initialization for the weights, which is specifically designed for networks with ReLU activations by maintaining an appropriate variance of activations throughout the network. This initialization scheme helps prevent the problem of exploding or vanishing gradients during the initial stages of training, facilitating faster convergence.

The Improved IKNet [6] uses a more compact structure with hidden dimensions [128, 64], relying on the enhanced representational power of residual connections rather than raw network width. This architecture maintains the same dropout rate of 0.1 as the Original IKNet[5] but applies it within the ResidualBlocks after each batch normalization and activation.

### 3.1.3 Focused IKNet Model

The Focused IKNet [6] architecture represents a significant departure from the previous models by explicitly modeling the distinct nature of position and orientation components in the inverse kinematics problem. Rather than processing the entire pose vector through a unified pathway, this architecture employs specialized processing branches that handle position and orientation separately before combining their extracted features.

As detailed in (4), the Focused IKNet [6] architecture consists of three main components below. In (4), h is position and orientation feature representations, W is branch-specific weight matrices, b is bias vectors, $h_1$, $h_2$ are subsequent hidden layer outputs and y is final joint angle prediction.

$$\begin{cases} x_{pos} &= x_{1:3} \quad \text{(position coordinates)} \\ x_{orient} &= x_{4:7} \quad \text{(orientation quaternion)} \\ h_{pos} &= \text{Dropout}(\text{ReLU}(W_{pos}x_{pos} + b_{pos}), p) \\ h_{orient} &= \text{Dropout}(\text{ReLU}(W_{orient}x_{orient} + b_{orient}), p) \\ h_{combined} &= [h_{pos}; h_{orient}] \quad \text{(concatenation)} \\ h_1 &= \text{Dropout}(\text{ReLU}(W_1 h_{combined} + b_1), p) \\ h_2 &= \text{ReLU}(W_2 h_1 + b_2) \\ y &= W_3 h_2 + b_3 \end{cases} \quad (4)$$

- A position branch that processes the 3D coordinates $(x, y, z)$ through a dedicated neural pathway, extracting position-specific features
- An orientation branch that handles the quaternion $(q_x, q_y, q_z, q_w)$ through a separate neural pathway, capturing orientation-specific patterns.
- A combine---d processing stage that integrates the features from both branches to predict joint angles, learning the complex interactions between position and orientation.

This architectural design is motivated by the observation that position, and orientation components often influence joint angles through different kinematic principles. Position primarily affects the reaching behavior of the manipulator, while orientation determines the wrist configuration. By providing dedicated processing paths, the network can develop specialized feature extractors for each component, potentially leading to more effective representation learning.

The Focused IKNet [6] employs a lower dropout rate of 0.05 compared to the 0.1 used in the other architectures. This reduction reflects the architecture's more structured approach to feature extraction, where the explicit separation of concerns reduces the risk of overfitting through architectural constraints rather than relying heavily on regularization techniques.

Both position and orientation branches use a hidden dimension of 64, which are then concatenated to form a 128-dimensional combined representation. This combined representation undergoes further processing through two fully connected layers before the final joint angle prediction. This progressive refinement allows the network to capture complex interactions between position and orientation features while maintaining computational efficiency.

### 3.1.4 Training procedure and optimization

All three network architectures were trained using a consistent procedure to ensure fair comparison. We employed the Adam optimizer with a learning rate of 1e-3 and weight decay of 1e-5 for regularization. The training process used a batch size of 128 samples and incorporated early stopping based on validation loss with a patience of 10 epochs. Additionally, we implemented learning rate reduction on plateau with a factor of 0.5 and patience of 5 epochs to adapt the optimization process as training progressed.

The loss function combined position error, orientation error, and joint limit penalties, as defined by (5). In (5), $E_{pos}$ is the mean squared error between predicted and target end-effector positions, $E_{orient}$ is the angular error between predicted and target orientations, $\theta_{i,max}$ represents the maximum allowed angle for joint i, w is weight and n is the number of joints.

$$L = w_{pos} \cdot E_{pos} + w_{orient} \cdot E_{orient} + w_{limit} \cdot \sum_{i=1}^{n} max(0, |\theta_{i,max}|)^2 \quad (5)$$

Where $E_{pos}$ is the Euclidean distance between target and achieved end-effector positions, $E_{orient}$ is the angular distance between target and achieved orientations, and the third term penalizes joint angles that exceed their mechanical limits. The weights $w_{pos}$, $w_{orient}$, and $w_{limit}$ were set to 1.0, 0.5, and 0.2 respectively, reflecting the relative importance of position accuracy, orientation alignment, and joint feasibility.

This comprehensive training approach ensures that the models not only learn accurate inverse kinematics mappings but also generate solutions that respect the physical constraints of the robotic manipulator.

### 3.2 XAI analysis

Understanding the decision-making process of neural networks is crucial for deploying them in safety-critical applications such as robotic manipulation. We implement a multi-faceted explainable AI (XAI) approach to analyze our IKNet models, combining state-of-the-art techniques to provide comprehensive insights into model behavior. This section details the methodologies employed to make the neural inverse kinematics models more transparent and interpretable.

### 3.2.1 SHAP analysis

This study employs SHapley Additive exPlanations (SHAP) as our primary tool for understanding feature importance. SHAP values provide unified measures of feature attribution based on cooperative game theory principles, offering several key advantages over other explanation methods mentioned below.

- Consistency: SHAP values satisfy the desirable property that when a model is changed to make a feature more important, its attribution does not decrease.
- Local accuracy: The sum of feature attributions equals the output of the model, providing mathematical precision.
- Missingness: Features with no impact on the prediction receive zero attribution.

The SHAP analysis implementation, detailed in Algorithm 1, follows a systematic process for each IKNet model.

---
**Algorithm 1** SHAP Analysis for IKNet Models
---
**Data:** model, background_data, test_data, feature_names
**Result:** SHAP values and expected values
Create CPU copy of model for analysis
Define prediction function for model outputs
Initialize KernelExplainer with prediction function
Compute SHAP values for test data
**return** *SHAP values, expected values*

---

First, we create a CPU copy of the model to ensure compatibility with the SHAP library. We then define a prediction function that takes input features and returns model outputs, which serves as the interface between the SHAP explainer and our model. We create a background dataset of 50 representative samples that serve as the baseline for feature attribution. This background set establishes the reference point from which feature contributions are measured.

Using the Kernel Explainer from the SHAP library, we compute SHAP values for a test dataset of 20 samples, limiting the sample size to maintain computational feasibility while ensuring statistically meaningful results. The computational core involves calculating the SHAP values $\emptyset_i$ for each feature $i$ according to (6), which quantifies the marginal contribution of each feature across all possible feature subsets. In (6), $\varphi_i(v)$ is the SHAP value for feature i, M is the set of all input features, S represents any subset of features not including feature i and v(S) is the model prediction using only features in S.

$$\phi_i(v) = \sum_{S \subseteq \mathcal{M} \setminus \{i\}} \frac{|S|!(|\mathcal{M}| - |S| - 1)!}{|\mathcal{M}|!} [v(S \cup \{i\}) - v(S)] \quad (6)$$

For neural networks, exact computation of SHAP values would require $2^7 = 128$ model evaluations per sample (for our 7-dimensional input), which becomes computationally prohibitive for large models and datasets. Therefore, we approximate them using the Kernel SHAP algorithm, which employs a weighted linear regression to estimate SHAP values with fewer model evaluations. This approximation achieves a balance between computational efficiency and explanation accuracy.

The SHAP analysis reveals which components of the end-effector pose (position coordinates or orientation quaternion) most significantly influence each joint angle prediction. This information provides crucial insights into the internal reasoning of the model, highlighting potential biases, unexpected dependencies, or counter-intuitive patterns.

### 3.2.2 Custom feature importance analysis

To complement the SHAP analysis and provide a different perspective on feature importance, this study implements a perturbation-based feature importance analysis. This approach, described in Algorithm 2, offers an intuitive and direct measure of feature influence based on the impact of feature perturbation on model predictions.

---
**Algorithm 2** Custom Feature Importance Analysis
---
**Data:** model, X_sample, feature_names, joint_names
**Result:** Importance scores for features
**for** *each joint in joint_names* **do**
  Define prediction function for joint
  Get base predictions on original data
  **for** *each feature in feature_names* **do**
    Create dataset with feature randomly permuted
    Calculate importance as mean absolute prediction difference
  Visualize feature importance for current joint
**return** *feature importance results*

---

The procedure begins by defining a prediction function specific to each joint angle output, allowing us to analyze the feature importance patterns for each joint independently. For each of the seven input features (three position coordinates and four quaternion components), there are some steps below to do.

- Create a copy of the input dataset with the target feature's values randomly permuted across samples.
- Generate predictions using the perturbed dataset.
- Calculate the mean absolute difference between predictions on the original and perturbed datasets.
- Use this difference as a measure of the feature's importance.

This technique provides an intuitive interpretation of feature importance based on the direct impact of feature disruption on model outputs. Mathematically, the importance score $I_i$ for feature $i$ is calculated according to (7), which measures the expected absolute difference in predictions when the feature is permuted while keeping all other features unchanged. In (7), $I_i$ is the importance score for feature I, f(X) is the model prediction on original dataset X and E[·] denotes the expected value over the dataset.

$$I_i = \mathbb{E}[|f(X) - f(X^i)|] \qquad (7)$$

The custom importance analysis offers several advantages as a complementary method to SHAP that mentioned below.
- It provides a model-agnostic approach that requires no assumptions about model structure.
- Computational complexity scales linearly with the number of features.
- The method has an intuitive interpretation as "how much does the prediction change when this feature is randomized?"
- It can be applied to any type of model without modification.

For each joint angle, we generate visualizations showing the relative importance of each input dimension. These visualizations help identify patterns such as which joints rely more heavily on position versus orientation information, potentially revealing insights about the kinematic structure that the model has learned.

### 3.2.3 Partial dependence analysis
While feature importance measures provide valuable insights into which input dimensions influence the model's predictions, they do not reveal how these features affect the outputs. To address this limitation, we conduct a partial dependence analysis as formalized in Algorithm 3, which reveals the marginal effect of each feature on the model's predictions after accounting for the average influence of all other features.

---
**Algorithm 3** Partial Dependence Analysis
**Data:** model, X_sample, feature_names, joint_names
**Result:** Partial dependence plots and values
for *each joint in joint_names* do
  for *each feature in feature_names* do
    Create grid of values spanning feature range
    for *each grid point* do
      Replace feature values with grid point value
      Calculate average prediction across modified dataset
    Generate partial dependence plot
**return** *partial dependence results*

---

The steps below are to combine each input feature and joint angle output.
- Create a grid of 20 values spanning the range of the feature based on its distribution in the dataset.
- For each grid point, replace the feature's value with the grid point value in all input samples.
- Generate predictions for these modified inputs and calculate the average prediction across all samples.
- Plot the relationship between the feature value and the average prediction.

The partial dependence plot (PDP) for feature $i$ is calculated according to (8), where $x_i$ is a specific value of feature $i$, $x_c$ represents all other features, and the expectation is taken over the marginal distribution of $x_c$ In practice, this expectation is approximated by averaging over the empirical distribution of $x_c$ in the dataset. In (8), PD$_i$(x$_i$) is the partial dependence of the model on feature i at value x$_i$, E$_{xc}$[·] is the expectation over the marginal distribution of all other features, xc represents all features except feature I and f is the model prediction given specific values.

$$\mathrm{PD}_i(x_i) = \mathbb{E}_{x_C}[f(x_i, x_C)] \qquad (8)$$

These plots reveal important patterns in how the model uses each input dimension as mentioned below.
- Monotonic relationships indicate consistent directional influence (e.g., increasing a coordinate consistently increases a particular joint angle).
- Non-monotonic relationships reveal complex dependencies (e.g., optimal values or threshold effects).
- Flat regions suggest insensitivity to the feature within certain ranges.
- Steep regions highlight areas of high sensitivity where small input changes cause large output changes.

By analyzing these patterns across different models, we can identify how architectural differences impact the functional relationships learned by each network. For example, we can determine whether the specialized branches in Focused IKNet [6] lead to different response patterns for position versus orientation features compared to the unified processing in Original IKNet[5].

### 3.2.4 Feature interaction analysis
Neural networks excel at capturing complex interactions between input features that cannot be understood through individual feature importance alone. Our feature interaction analysis, outlined in Algorithm 4, systematically explores how pairs of important features jointly influence model predictions, revealing nonlinear interactions that might be missed by single-feature analyses.

---
**Algorithm 4** Feature Interaction Analysis
**Data:** model, X_sample, importance_scores, feature_names, joint_names
**Result:** Feature interaction strengths
for *each joint in joint_names* do
  Get top features based on importance scores
  for *each pair of top features (feat1, feat2)* do
    Create 2D grid for both features
    for *each grid combination (val1, val2)* do
      Create dataset with feature pair set to grid values
      Store average predictions in interaction matrix
    Visualize interaction with heatmap
**return** *interaction matrices*

---

Based on the feature importance results from previous analyses, identifying the top three features for each joint angle output. For each pair of these important features, there are some steps that are mentioned below.
- Create a 2D grid spanning the range of each feature pair (10×10 grid points).

- For each grid point combination, set the values of the feature pair to the grid point values while maintaining other features at their original values.
- Compute the model's predictions for each modified input and calculate the average prediction.
- Visualize the interaction using a 2D heatmap where color intensity represents the average predicted joint angle.

The interaction strength between features *i* and *j* is quantified according to (9), which measures the portion of the model's output variation that cannot be explained by the additive effects of individual features. High interaction strength indicates that the features work together in a way that produces effects beyond their individual contributions.

In (9), $I$ represent to interaction strength and f represents to joint effect of features.

$$I_{i,j} = \sum_{x_i, x_j} |f_{i,j}(x_i, x_j) - f_i(x_i) - f_j(x_j) + f_0| \quad (9)$$

This analysis reveals complex relationships that are shown below.
- Conditional dependencies, where the effect of one feature depends on the value of another.
- Synergistic interactions, where features amplify each other's effects.
- Antagonistic interactions, where features mitigate each other's effects.
- Threshold interactions, where both features must exceed certain values to influence the output.

The feature interaction analysis is particularly valuable for understanding the behavior of more complex architectures like Focused IKNet [6], where the explicit separation of position and orientation processing might lead to different interaction patterns compared to the unified processing in Original IKNet[5] and Improved IKNet [6].

### 3.2.5 XAI visualization

To facilitate comprehensive interpretation of the XAI results, this study develops a visualization framework that presents the various analyses in an integrated and accessible manner. For each model and joint angle combination, the results below are generated.
- Bar charts show the relative importance of each input dimension from both SHAP and custom importance analyses.
- Line plots display the partial dependence of joint angles on each input feature.
- Heatmaps illustrate feature interactions between pairs of important features.
- Combined visualizations that connect feature importance to actual arm configurations.

These visualizations are organized hierarchically, allowing for both high-level comparison across models and detailed examination of specific joints or features. Additionally, we implement interactive components that enable exploration of the relationship between explanations and model predictions for different input scenarios.

The visualization framework serves not only as an analysis tool but also as a communication medium that makes the complex behavior of neural inverse kinematics models accessible to robotics practitioners who may not have expertise in deep learning or explainable AI techniques.

### 3.2.6 InterpretML analysis

To complement the SHAP analysis and provide a more comprehensive understanding of model behavior, this study implements additional explainability techniques inspired by the InterpretML framework. This approach, detailed in Algorithm 5, provides a multi-faceted view of feature importance and interactions using both model-agnostic and model-specific techniques.

The InterpretML analysis begins with data sampling to ensure computational efficiency while maintaining statistical validity. The analysis is limited to 200 samples, which balances the need for robust results against computational constraints.

For feature importance analysis, this study implements a custom permutation-based approach that mentioned below.
- Measures the impact of randomly shuffling each feature on model predictions.
- Quantifies importance as the mean absolute difference in predictions.
- Generates feature importance distributions for each joint independently.

---
**Algorithm 5** InterpretML Analysis

**Require:** model, dataset, feature_names, joint_names, output_dir
**Ensure:** Dictionary of interpretability results
1: Sample data for analysis (up to 200 samples)
2: Create predict function for model
3: // Custom Feature Importance Analysis
4: **for** each joint in joint_names **do**
5:   **for** each feature in feature_names **do**
6:     Create perturbed dataset with shuffled feature values
7:     Calculate importance as mean absolute prediction difference
8:     Visualize feature importance
9:   **end for**
10: **end for**
11: // Manual Partial Dependence Analysis
12: **for** each joint in joint_names **do**
13:   **for** each feature in feature_names **do**
14:     Create grid spanning feature range
15:     **for** each grid point **do**
16:       Replace feature values and calculate average prediction
17:       Store in partial dependence values
18:     **end for**
19:     Generate partial dependence plot
20:   **end for**
21: **end for**
22: // Feature Interaction Analysis
23: **for** each joint in joint_names **do**
24:   Identify top features by importance
25:   **for** each pair of top features **do**
26:     Create 2D grid for feature combinations
27:     Calculate and visualize interaction effects
28:   **end for**
29: **end for**
30: Create consolidated visualization of all analyses
31: **return** interpretability results =0

---

This technique provides an intuitive interpretation of feature influence based on the direct impact of feature disruption, which is particularly valuable for understanding the relationship between pose components and specific joint angles.

The manual partial dependence analysis explores how each input dimension affects the predicted joint angles across its range of values. This reveals nonlinear relationships and response thresholds that might not be apparent from aggregate importance measures. By visualizing these relationships, some results could be identified as mentioned below.
- Monotonic relationships indicating consistent directional influence.
- Plateau regions where the model is insensitive to the feature.
- Critical threshold points where small input changes have large effects.
- Discontinuities that might indicate model instabilities or dataset artifacts.

The feature interaction analysis that has proposed focuses on the most important features identified in the previous steps, examining how pairs of these features jointly influence model predictions. This is particularly relevant for inverse kinematics, where the interactions between position and orientation components often have complex effects on joint angles.

The consolidated visualization brings together results from all analyses to provide a holistic view of each model's interpretation patterns. This integration helps identify consistent patterns across analysis techniques, increasing confidence in the interpretations, as well as discrepancies that might warrant further investigation.

Unlike standard model-agnostic interpretability approaches that treat all outputs equally, what this study has implemented specifically analyzes each joint output independently. This joint-specific approach provides more detailed insights into how different parts of the kinematic chain respond to input features, revealing specialized joint behaviors that might be obscured in a unified analysis.

The InterpretML analysis serves as both a validation mechanism for the SHAP results and a source of additional insights that might not be captured by any single explainability technique. By triangulating findings across multiple explanation methods, the robustness and reliability have increased of this study interpretations of model behavior.

## 3.3 Obstacle avoidance

A critical aspect of robotic manipulation is the ability to navigate environments with obstacles safely. This study has proposed a comprehensive obstacle avoidance evaluation framework to assess how well our IKNet models implicitly learn to generate joint configurations that avoid collisions while accurately reaching target poses. This section details our methodology for simulating, detecting, and analyzing obstacle avoidance behavior.

### 3.3.1 Forward kinematics implementation

To evaluate the arm configurations produced by our models, this study implements forward kinematics calculations as described in Algorithm 6. This algorithm transforms the predicted joint angles into the positions of each joint and the end-effector in either 2D or 3D space, providing a geometric representation of the manipulator that can be used for collision detection and visualization.

---
**Algorithm 6** Forward Kinematics
**Require:** joint_angles, link_lengths, add_z
**Ensure:** Joint positions
1: positions = add_z ? [(0,0,0)] : [(0,0)]
2: cumulative_angle = 0
3: **for** i, angle in enumerate(joint_angles) **do**
4:    cumulative_angle += angle
5:    prev_pos = positions[-1]
6:    x = prev_pos[0] + link_lengths[i] * cos(cumulative_angle)
7:    y = prev_pos[1] + link_lengths[i] * sin(cumulative_angle)
8:    **if** add_z **then**
9:      z = prev_pos[2] + link_lengths[i] * 0.1 * (i+1)/len(link_lengths)
10:      positions.append((x,y,z))
11:    **else**
12:      positions.append((x,y))
13:    **end if**
14: **end for**
15: **return** positions =0
---

For a manipulator with $n$ joints and link lengths $\{l_1, l_2, l_3, l_4\}$, the position of joint $i$ is calculated using cumulative angles to account for the relative positioning of each joint. For 2D calculations, shown as (10), $(x,y)$ coordinates of each joint based on the cumulative angle and link length are calculated. For 3D calculations, shown as (11), a z-coordinate that increases slightly with each joint to create a more realistic visualization with depth perspective are added. In (10), $\theta_{cum,i}$ is the cumulative angle up to joint i, $x_i$, $y_i$ are the 2D coordinates of joint I and $l_j$ is the length of link j. In (11), all the definitions are like (10) and $z_i$ is vertical offset for 3D visualization.

$$\begin{cases} \theta_{cum,i} = \sum_{j=1}^{i} \theta_j \\ x_i = x_{i-1} + l_i \cos(\theta_{cum,i}) \\ y_i = y_{i-1} + l_i \sin(\theta_{cum,i}) \end{cases} \quad (10)$$

$$\begin{cases} \theta_{cum,i} = \sum_{j=1}^{i} \theta_j \\ x_i = x_{i-1} + l_i \cos(\theta_{cum,i}) \\ y_i = y_{i-1} + l_i \sin(\theta_{cum,i}) \\ z_i = z_{i-1} + l_i \cdot 0.1 \cdot \frac{i}{n} \end{cases} \quad (11)$$

The implementation of this study uses link lengths [2.0, 1.8, 1.5, 1.0] representing a 4-DOF robotic manipulator with decreasing segment lengths, which is a common configuration in practical robot designs. The forward kinematics module is implemented in a vectorized manner using NumPy to enable efficient processing of multiple arm configurations simultaneously.

The forward kinematics calculation serves several purposes in our evaluation framework as shown below.
- Providing the end-effector position for measuring target reaching accuracy
- Generated the complete arm configuration for collision detection

- Enabled visualization of the model's predictions in physical space.
- Facilitated comparative analysis of different models in the same scenario.

### 3.3.2 Collision detection algorithm

Once the arm configuration is determined through forward kinematics, we perform collision detection between the arm segments and obstacles in the environment. Algorithm 6 details our collision detection approach, which calculates the minimum clearance between any arm segment and any obstacle.

For each combination of obstacle and arm segment, the steps are mentioned as below.
- Extract the coordinates of the segment endpoints $(x_1, y_1)$ and $(x_2, y_2)$.
- Calculate the line vector $v$ and its length from these endpoints.
- Compute the projection parameter $t$ that determines the closest point on the line to the obstacle center.
- Constrain $t$ to the range [0,1] to ensure the closest point lies on the segment rather than the extended line.
- Calculate the closest point on the segment to the obstacle center using this constrained parameter.
- Compute the distance between this point and the obstacle center.
- Determine the clearance as the difference between this distance and the obstacle radius.

The closest point on a line segment to a point is calculated according to (12), which involves a projection calculation followed by parameter clamping to ensure the result lies on the segment.

$$\begin{cases} \mathbf{v} = \mathbf{x}_2 - \mathbf{x}_1 & \text{(segment vector)} \\ t = \frac{(\mathbf{o}-\mathbf{x}_1)\cdot\mathbf{v}}{|\mathbf{v}|^2} & \text{(projection parameter)} \\ t' = \max(0, \min(1, t)) & \text{(constrained parameter)} \\ \mathbf{p} = \mathbf{x}_1 + t'\mathbf{v} & \text{(closest point)} \end{cases} \quad (12)$$

Collision detection algorithm tracks several key metrics as mentioned below.
- Minimum clearance across all segment-obstacle pairs.
- Which arm segment has the smallest clearance (critical segment).
- Which obstacle is closest to collision (critical obstacle).
- Whether any collision occurs (clearance $\leq 0$).

This detailed collision analysis provides insights not just into whether the arm configuration avoids obstacles, but also how close it comes to collision and which parts of the arm are most critical for obstacle avoidance. This information is valuable for understanding how different IKNet architectures approach the implicit obstacle avoidance problem.

### 3.3.3 Step-by-step obstacle avoidance analysis

To gain deeper insights into how the different models handle obstacle avoidance, this study implements a detailed step-by-step analysis procedure as outlined in Algorithm 7. This comprehensive analysis breaks down the obstacle avoidance process into discrete steps, providing fine-grained information about each model's behavior.

The step-by-step analysis follows this procedure shown below.
- Model prediction: Provide the target end-effector pose to the IKNet model and obtain the predicted joint angles. For consistency across models, use the same target pose format: 3D position coordinates and a default orientation quaternion representing no rotation.
- Forward kinematics: Calculate the complete arm configuration using the predicted joint angles and the specified link lengths, generating both 2D and 3D representations for different analysis purposes.
- Per-obstacle collision checking: For each obstacle in the scenario, performed detailed collision detection with each arm segment, recording the minimum clearance, whether a collision occurs, and which segment has the smallest clearance to the obstacle.
- End-effector error calculation: Compute the Euclidean distance between the achieved end-effector position (from forward kinematics) and the target position, measuring the positioning accuracy independent of collision avoidance.
- Critical area identification: Based on the collision results, identify the most critical segment-obstacle pair (with minimum clearance) and perform detailed geometric analysis of this critical area, including visualization of the closest point and distance clearance.

---
**Algorithm 7** Collision Detection
---
**Require:** arm_positions, obstacle_positions, obstacle_radii
**Ensure:** Collision status, clearance, critical segment
1: min_clearance = infinity
2: collision_detected = false
3: collision_segment = null
4: collision_obstacle = null
5: **for** obs_idx, (obs_pos, obs_radius) in enumerate(obstacles) **do**
6:   **for** segment_idx in range(len(arm_positions) - 1) **do**
7:     Calculate line vector and length from segment endpoints
8:     **if** line_len ¿ 0 **then**
9:       Calculate closest point on segment to obstacle
10:      distance = distance between closest point and obstacle center
11:      clearance = distance - obs_radius
12:      **if** clearance ¡ min_clearance **then**
13:        min_clearance = clearance
14:        collision_segment = segment_idx
15:        collision_obstacle = obs_idx
16:      **end if**
17:      **if** distance ¡= obs_radius **then**
18:        collision_detected = true
19:      **end if**
20:     **end if**
21:   **end for**
22: **end for**
23: **return** collision_detected, min_clearance, collision_segment, collision_obstacle =0
---

This step-by-step approach provides several advantages over a simple binary collision check as shown below.
- Quantifies the safety margin in non-collision cases.
- Identify which part of the solution needs improvement in collision cases.

- Reveal the trade-offs between obstacle avoidance and target reaching accuracy.
- Enable detailed visualization and comparison of different models' approaches.

For each model and scenario, this study generates comprehensive visualizations showing the arm configuration, obstacles, critical areas, and performance metrics. These visualizations facilitate both quantitative assessment and intuitive understanding of each model's obstacle avoidance capabilities.

### 3.3.4 Multiple obstacle scenario generation

To ensure a robust evaluation, this study generates diverse obstacle avoidance scenarios as described in Algorithm 8. Each scenario contains multiple obstacles with random positions, sizes, and heights, along with a target end-effector position that is reachable without necessarily colliding with obstacles.

---
**Algorithm 8** Collision Detection
---
**Require:** arm_positions, obstacle_positions, obstacle_radii
**Ensure:** Collision status, clearance, critical segment
1: min_clearance = infinity
2: collision_detected = false
3: collision_segment = null
4: collision_obstacle = null
5: **for** obs_idx, (obs_pos, obs_radius) in enumerate(obstacles) **do**
6:   **for** segment_idx in range(len(arm_positions) - 1) **do**
7:     Calculate line vector and length from segment endpoints
8:     **if** line_len > 0 **then**
9:       Calculate closest point on segment to obstacle
10:       distance = distance between closest point and obstacle center
11:       clearance = distance - obs_radius
12:       **if** clearance < min_clearance **then**
13:         min_clearance = clearance
14:         collision_segment = segment_idx
15:         collision_obstacle = obs_idx
16:       **end if**
17:       **if** distance <= obs_radius **then**
18:         collision_detected = true
19:       **end if**
20:     **end if**
21:   **end for**
22: **end for**
23: **return** collision_detected, min_clearance, collision_segment, collision_obstacle =0

---

The scenario generation process follows the steps mentioned below.

- Determine the number of obstacles for the scenario (randomly selected between 2 and 5).
- For each obstacle:
  - Generate a random angle and distance from the origin to determine the obstacle position.
  - Ensure obstacles are not too close to the manipulator base (minimum distance of 1.5 units).
  - Assign a random radius between 0.3 and 0.8 units.
  - Set a random height between 0.5 and 2.0 units for 3D visualization.
- Generate a target position:
  - Select a random angle and distance from the origin.
  - Ensure the target is within a reasonable reach of the manipulator (distance between 3.0 and 5.0 units).
  - Verify that the target does not collide with any obstacle (with a 0.3-unit buffer).
  - If a collision is detected, repeat the target generation process.

This approach creates challenging but solvable scenarios that test the models' ability to navigate complex environments. The scenarios vary in difficulty based on the elements mentioned below.

- The number of obstacles (more obstacles create more constrained environments).
- The placement of obstacles is related to the direct path to the target.
- The size of obstacles (larger radii create narrower passages).
- The relative position of the target (requiring different arm configurations).

Ten diverse scenarios are generated for our primary evaluation, providing a consistent test suite across all models. This standardization ensures fair comparison while covering a range of obstacle arrangements and target positions that reflect different real-world manipulation challenges.

### 3.3.5 Dynamic obstacle analysis

Beyond static obstacle scenarios, how the models perform with moving obstacles to assess their robustness to changing environments are also evaluated. A dynamic obstacle analysis that simulates obstacles moving along predefined trajectories while the target position remains fixed is also implemented.

Each model is evaluated by the metrics mentioned below.

- Response to gradually approaching obstacles.
- Behavior when passages between obstacles narrow over time.
- Recovery capability when obstacles suddenly appear in the path.
- Stability of solutions under minor perturbations to obstacle positions.

This dynamic analysis provides insights into how well the inverse kinematics models generalize to non-static environments, an important consideration for real-world robotic applications where obstacle configurations may change during operation.

### 3.4 Comparative evaluation

To comprehensively assess the relative strengths and weaknesses of our three IKNet architectures, a systematic comparative evaluation methodology is implemented. This approach integrates results from both XAI analysis and obstacle avoidance simulation to provide a holistic understanding of model performance and behavior.

### 3.4.1 Feature importance comparison

A comparative analysis of feature importance across models and joints using a heatmap visualization approach is implemented as detailed in Algorithm 9. This visualization reveals patterns in how different architectures utilize input features and how this utilization varies across joint outputs.

For each model, an important matrix of shape (joints × features) where each cell represents the normalized importance of a specific feature for a specific joint angle are constructed. These important values from the SHAP analysis results are extracted, handling the different data structures that may arise based on the SHAP output format.

The comparative analysis is quantified using (13), which normalizes feature importance scores for each model to enable direct comparison. This normalization ensures that differences in the magnitude of importance scores do not obscure the relative importance patterns across models. In (13), $\hat{I}_{m,i}$ is the normalized importance, $I_{m,i}$ is the raw importance score.

---

**Algorithm 9** Obstacle Scenario Generation
---
**Require:** num_scenarios, max_obstacles
**Ensure:** Random obstacle scenarios
1: Initialize empty scenarios list
2: **for** s = 1 to num_scenarios **do**
3:   Generate random number of obstacles (2 to max_obstacles)
4:   **for** each obstacle **do**
5:     Generate random position, radius, and height
6:     Add obstacle to scenario
7:   **end for**
8:   **repeat**
9:     Generate random target position
10:    collision = false
11:    **for** each obstacle in obstacles **do**
12:      **if** distance ¡ obstacle.radius + 0.3 **then**
13:        collision = true
14:        break
15:      **end if**
16:    **end for**
17:  **until** not collision
18:  Add scenario to scenarios list
19: **end for**
20: **return** scenarios =0

$$\bar{I}_{m,i} = \frac{I_{m,i}}{\max_j I_{m,j}} \tag{13}$$

The heatmap visualization can highlight several key aspects that are shown below.
- Which models rely more heavily on position versus orientation components.
- How feature importance patterns differ across joints for each model.
- Whether models exhibit consistent feature utilization patterns or highly variable dependencies.
- Which architectural designs lead to more focused or distributed feature utilization.

By comparing these patterns across models, the architectural differences influence the models' internal representations and decision processes can be identified. For instance, we can determine whether the specialized branches in Focused IKNet [6] lead to more distinct separation of position and orientation influences compared to the unified processing in Original IKNet[5] and Improved IKNet [6].

### 3.4.2 Obstacle avoidance performance metrics

To evaluate obstacle avoidance performance, a set of metrics computed across multiple scenarios is defined as outlined in Algorithm 10. These metrics provide a multi-faceted view of each model's ability to generate joint configurations that both avoid obstacles and accurately reach target positions.

For each model, the following metrics mentioned can be computed.
- Minimum clearance: The smallest distance between any arm segment and any obstacle across all scenarios. This metric captures the model's margin of safety during obstacle avoidance, with higher values indicating more conservative solutions that maintain greater distance from obstacles.
- Target position error: The Euclidean distance between the achieved end-effector position and the target position. This metric measures the accuracy of the inverse kinematics solution independent of obstacle considerations, with lower values indicating better target reaching precision.
- Collision rate: The percentage of scenarios where the arm configuration results in a collision with at least one obstacle. This binary metric provides a high-level assessment of obstacle avoidance success, with lower rates indicating better performance.
- Critical segment identification: Analysis of which arm segments most frequently have the smallest clearance to obstacles. This metric reveals whether certain parts of the arm consistently pose greater collision risks, providing insights into the models' obstacle avoidance strategies.

---

**Algorithm 10** Feature Importance Heatmap
---
**Require:** shap_values_dict, feature_names, joint_names
**Ensure:** Heatmap visualization
1: Create figure with subplots for each model
2: **for** each model, shap_data in shap_values_dict **do**
3:   Initialize importance matrix of shape (joints, features)
4:   Extract importance values from SHAP data
5:   Create heatmap visualization with annotated values
6: **end for**
7: Add colorbar and save figure
8: **return** visualization figure =0

These metrics are aggregated across all test scenarios to provide robust performance statistics that account for diverse obstacle arrangements and target positions. The aggregation process calculates mean values, standard deviations, and confidence intervals for each metric, enabling statistically sound comparisons between models.

### 3.4.3 Multi-model comparison visualization

To facilitate intuitive understanding of comparative performance, a multi-model comparison visualization is implemented as described in Algorithm 11. This visualization presents all models' solutions for the same scenario in a unified display, enabling direct visual comparison of their approaches and outcomes.

The visualization includes several components that are shown below.
- 2D overview: A top-down view showing the arm configurations from all models alongside obstacles and the target position. Each model's solution is rendered with a different color for easy identification, and segments are highlighted based on their clearance to obstacles.
- Performance metrics: Bar charts comparing quantitative metrics (clearance and target error) across models for the current scenario. These charts include value labels and visual indicators for collision status, providing an immediate assessment of relative performance.
- Summary table: A structured presentation of key metrics for each model, including minimum clearance, target error, and collision status. The table uses conditional formatting to highlight critical values, making it easy to identify the best-performing model for different criteria.

For scenarios with identified collisions or near-collisions, the visualization also includes a detailed closeup view of the critical area. This closeup highlights the specific segment-obstacle interaction that represents the minimum clearance, showing the exact geometry of the potential collision point.

This comprehensive visualization enables both qualitative assessment of the models' approaches to obstacle avoidance and quantitative comparison of their performance metrics. The intuitive presentation makes complex performance differences accessible even to non-experts, facilitating informed model selection for specific applications.

---

**Algorithm 11** Performance Metrics Calculation

---
**Require:** all_collision_results
**Ensure:** Aggregated metrics
1: Initialize results dictionary
2: **for** each model, results in all_collision_results **do**
3:   Calculate average clearance across scenarios
4:   Calculate average target error across scenarios
5:   Count collision occurrences
6:   Calculate collision rate percentage
7:   Store metrics in results dictionary
8: **end for**
9: **return** aggregated metrics =0

---

### 3.4.4 Comprehensive summary table

Finally, all analysis results are integrated into a comprehensive summary table generated according to Algorithm 12. This table consolidates findings from both XAI analysis and obstacle avoidance evaluation, providing a unified view of each model's characteristics and performance.

For each model, the summary table includes the elements mentioned below.
- Top features from SHAP analysis: The three most influential input dimensions for each model according to SHAP analysis, revealing which pose components drive the model's predictions.
- Top features from custom importance analysis: The three most important features identified by the perturbation-based method, providing an alternative perspective on feature influence.
- Obstacle avoidance metrics: Key performance statistics including average clearance, target error, and collision rate across all evaluation scenarios.
- Computational performance: Metrics such as inference time and memory usage that characterize the computational efficiency of each model.
- Overall strengths and weaknesses: A qualitative assessment of each model's distinctive advantages and limitations based on the combined analysis results.

---

**Algorithm 12** Model Comparison Visualization

---
**Require:** results, link_lengths, obstacles, target_position
**Ensure:** Comparative visualization
1: Create multi-panel figure for 2D view and metrics
2: Plot obstacles and target position
3: **for** each model, result in results **do**
4:   Plot arm configuration with color-coding
5:   Add annotations for clearance and positioning error
6: **end for**
7: Create comparative bar charts for key metrics
8: Add performance summary table with highlighting
9: **return** visualization figure =0

---

The summary table is complemented by visualizations comparing key metrics across models, providing an accessible overview of the comparative analysis results. This integrated view enables identification of correlations between explainability results and performance metrics, such as whether models that focus more on certain pose components achieve better obstacle avoidance or target accuracy.

## 4 Results and evaluation

In this section, we present a comprehensive analysis of the three IKNet variants: Original IKNet[5], Improved IKNet [6], and Focused IKNet [6]. Through quantitative and qualitative assessments, we examine their performance in obstacle avoidance tasks, identify the key factors influencing their decision-making processes, and evaluate their effectiveness across multiple scenarios. The results highlight significant differences in how these models approach obstacle avoidance challenges, revealing important insights into their underlying mechanisms and potential applications.

### 4.1 SHAP results

The SHAP analysis provides critical insights into how different joint variables influence the models' decision-making during obstacle avoidance tasks. SHAP values quantify the contribution of each feature to the prediction made by the model relative to the average prediction, offering a robust framework for model interpretability. Fig. 2-4 present the meaning of absolute SHAP values for each feature across the three IKNet models: Original IKNet[5], Improved IKNet [6], and Focused IKNet [6]. The analysis encompasses seven key features: three positional coordinates $(x, y, z)$ and four quaternion components $(q_x, q_y, q_z, q_w)$ that together define the complete kinematic state of the robotic system.

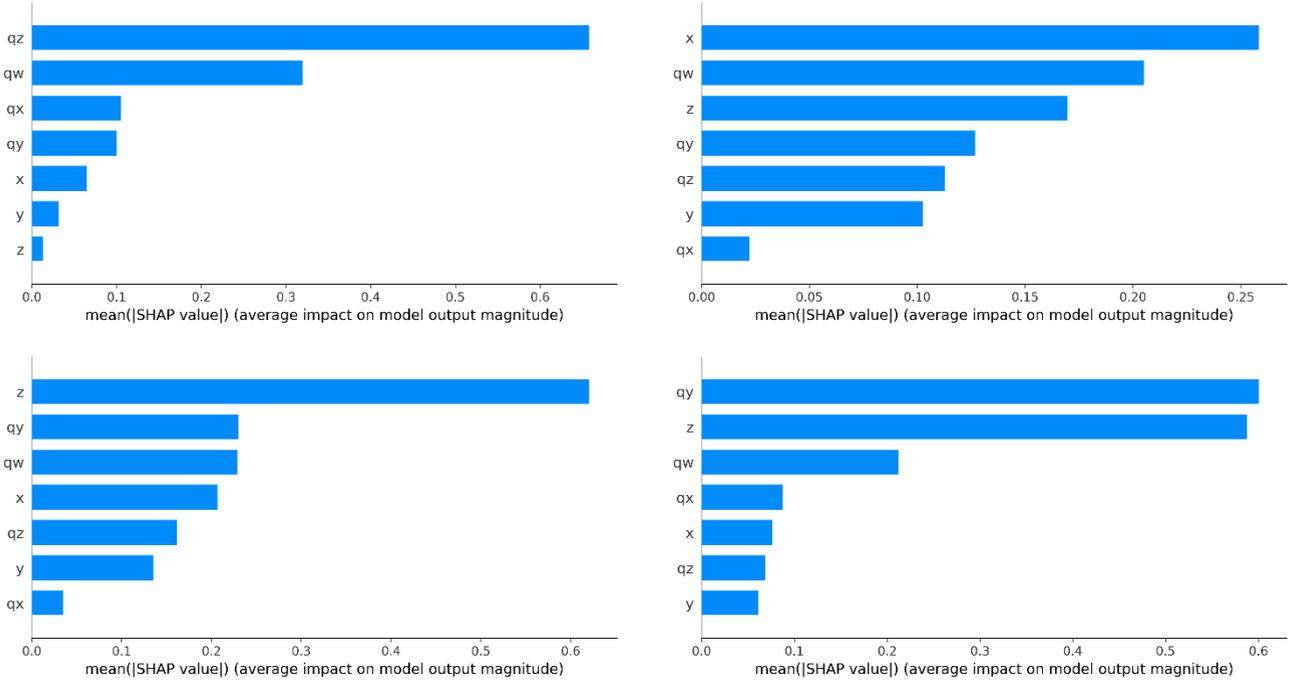

**Fig. 2.** SHAP value of Original IKNet model (row1: joint1 and joint2, row2: joint3 and joint 4)

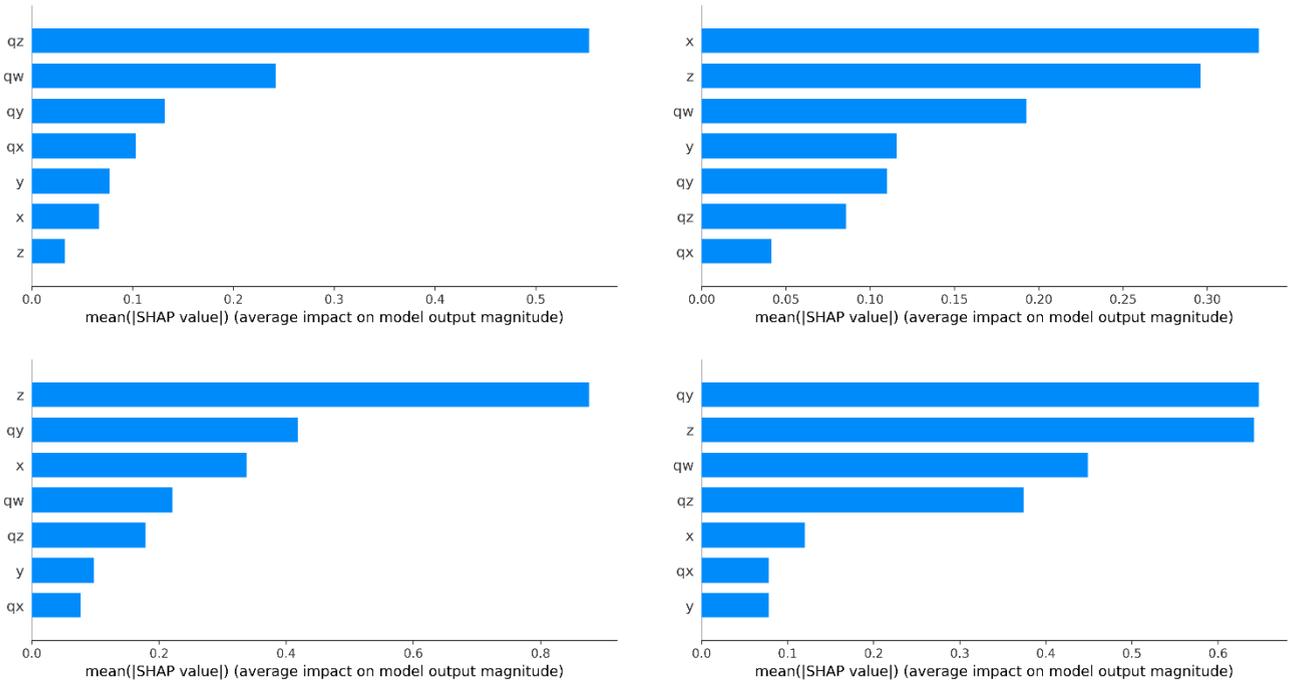

**Fig. 3.** SHAP value of Improved IKNet model (row1: joint1 and joint2, row2: joint3 and joint 4)

### 4.1.1 Original IKNET

In the Original IKNet[5] model (Fig. 2), the most influential features vary significantly across scenarios. For the first scenario, quaternion z (qz) demonstrates the highest impact with a mean |SHAP value| of approximately 0.6, followed by quaternion w (qw) at around 0.3. This indicates that rotation components along these axes have the greatest influence on the model's obstacle avoidance decisions. The predominance of quaternion components suggests that the Original IKNet[5] prioritizes orientation adjustments as its primary strategy for obstacle avoidance. Interestingly, the positional variables (x, y, z) show relatively minimal impact, suggesting that the original model relies more heavily on orientation information rather than absolute position. This behavior may be attributed to the model's training process, which potentially emphasized orientation-based solutions to obstacle avoidance challenges. The top-left plot in Fig. 2 clearly illustrates this pattern, with qz exhibiting a significantly larger bar compared to other features, followed by qw. The consistent pattern of higher importance for quaternion components across multiple scenarios (shown in the remaining plots) reinforces this orientation-centric approach of the Original IKNet[5].

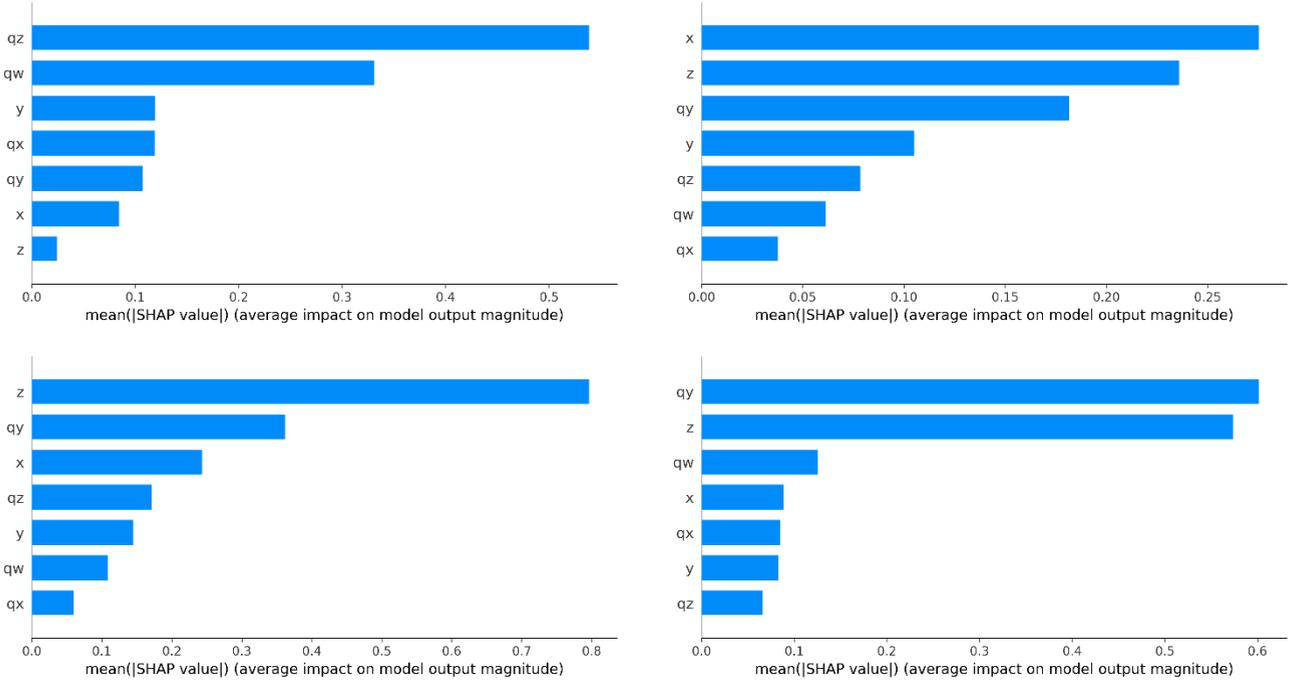

**Fig. 4.** SHAP value of Focused IKNet model (row1: joint1 and joint2, row2: joint3 and joint4)

### 4.1.2 Improved IKNet

The Improved IKNet [6] model (Fig. 3) exhibits a distinctly different pattern of feature importance. While quaternion z (qz) remains significant in some scenarios, the x position coordinate shows consistently higher importance across multiple test cases, with mean |SHAP values| reaching approximately 0.3. The top-right plot in Fig. 5 demonstrates this shift, with the x coordinate exhibiting a prominence comparable to that of quaternion components. Additionally, the z coordinate shows increased importance compared to the Original IKNet[5], particularly in the bottom-left plot where it ranks among the top three influential features. This significant shift suggests that the Improved IKNet [6] has developed a more balanced approach to obstacle avoidance, considering both positional and rotational information in its decision-making process. The integration of positional data into the model's strategy indicates a more comprehensive spatial awareness, potentially enabling more efficient navigation around obstacles while maintaining appropriate clearances.

### 4.1.3 Focused IKNet

The FocusedIKNet model (Fig. 4) demonstrates yet another distinct pattern, with quaternion y (qy) and z position showing the highest impact across scenarios. The mean |SHAP values| for these features reach approximately 0.6 and 0.5 respectively, as evident in the top-right plot of Fig. 6. The prominence of these specific features indicates that the FocusedIKNet places greater emphasis on particular rotational and vertical position information when navigating obstacles. This specialized approach suggests that the FocusedIKNet may have been trained to prioritize certain movement strategies, potentially targeting specific types of obstacle configurations or movement constraints. The bottom-left plot in Fig. 6 further illustrates this specialization, showing z as the most influential feature with a |SHAP value| considerably higher than other features. This consistent emphasis on z-axis positioning across scenarios indicates that vertical adjustments form a core component of the Focused IKNet [6]'s obstacle avoidance strategy.

### 4.1.4 Feature importance

Fig. 5 presents a comparative matrix of mean |SHAP values| across all joint positions for the three models, providing a consolidated view of feature importance across the entire kinematic chain. This matrix representation enables a direct comparison of how each model weighs different features for each joint, revealing fundamental differences in their obstacle avoidance strategies. The Original IKNet[5] shows high dependency on z-axis rotation for joints 3 and 4 (values of 0.62 and 0.59 respectively), with

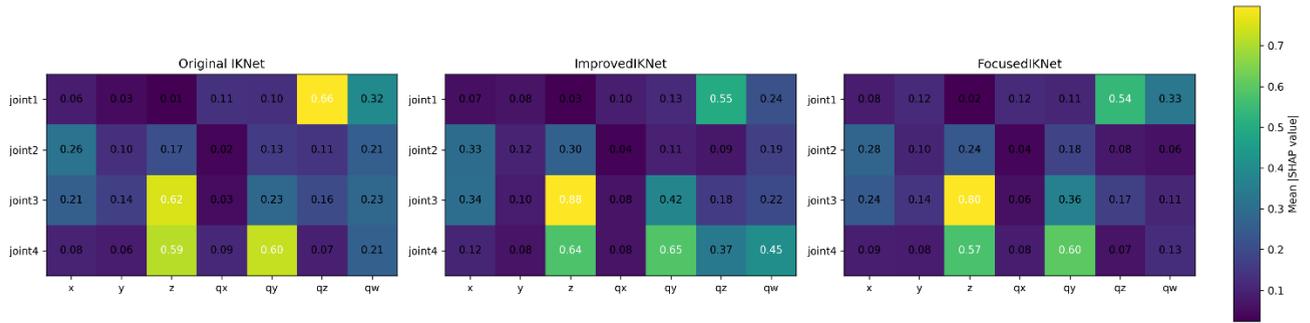

**Fig. 5.** Feature importance analysis by SHAP

relatively lower importance assigned to other features. This concentrated feature utilization suggests a specialized approach that heavily relies on specific joint rotations for obstacle avoidance. In contrast, the Improved IKNet making process. The Focused IKNet [6] exhibits the most specialized pattern, with particularly high importance assigned to z variables for joints 3 and 4 (0.60 and 0.57), suggesting a targeted approach that prioritizes specific movement patterns.

### 4.2 InterpretML results

The comprehensive analysis of the three IKNet variants reveals distinctive approaches to obstacle avoidance, with each model demonstrating unique patterns of feature utilization and decision-making strategies. These differences provide valuable insights into how neural network architectures can be optimized for robotic navigation tasks and highlight the relationship between feature importance and overall performance.

#### 4.2.1 Heat maps

The heat maps (Figs. 6-8) visualize the complex interaction patterns between features across the joint space, providing a complementary perspective to the bar charts. These visualizations represent the sensitivity of model outputs to variations in input features, with brighter colors (yellow-green) indicating higher sensitivity and darker colors (blue-purple) representing lower sensitivity.

In Fig. 6 for Original IKNet[5], the heat maps display distinct patterns of sensitivity that vary considerably across the feature space. Certain regions show concentrated brightness, particularly in areas corresponding to the interaction between z-position and quaternion components. This non-uniform distribution suggests that the Original IKNet[5] is highly sensitive to specific combinations of input values, potentially indicating a less generalized approach to obstacle avoidance that may perform well in certain configurations but less optimally in others.

The heat maps for Improved IKNet [6] in Fig. 7 reveal a more balanced distribution of sensitivity across the feature space. The gradient patterns appear more uniform with demonstrates more balanced utilization of features with significant weights distributed across z, qy, and qz variables. This distribution indicates a more comprehensive strategy that integrates multiple information sources into its decision-smoother transitions between regions of high and low sensitivity. This more distributed sensitivity profile aligns with the model's more balanced feature utilization observed in the SHAP analysis and likely contributes to its superior performance across diverse scenarios. The more uniform sensitivity indicates that the model responds more consistently to variations in inputs, enabling more robust obstacle avoidance behaviors.

For Focused IKNet [6] in Fig. 8, the heat maps show concentrated regions of high sensitivity with sharp transitions between high and low sensitivity areas. This pattern suggests that the model is highly responsive to specific input combinations but potentially less adaptive to variations outside these optimized regions. The concentrated sensitivity aligns with the specialized feature importance observed in the SHAP analysis, reinforcing the characterization of Focused IKNet [6] as employing a more targeted approach to obstacle avoidance.

#### 4.2.2 Feature Importance

Joint-specific analysis Fig. 9 provides deeper insights into the importance of each joint across the three models. These detailed breakdowns reveal how different joints contribute to the overall obstacle avoidance strategy, highlighting the specialized roles that each joint plays in the kinematic chain.

For the Original IKNet[5] in Fig. 9 top-left, joint3 and joint4 show particularly high sensitivity to the z-coordinate, with important scores of 0.413 and 0.435 respectively. This predominance suggests that the vertical positioning of these distal joints is critical to the model's obstacle avoidance strategy. In contrast, joint1 shows highest sensitivity to the z-coordinate (0.019) but with a much lower magnitude, while joint2 is most influenced by quaternion w (0.148). This hierarchical pattern indicates that the Original IKNet[5] implements a strategy where proximal joints (closer to the base) are more concerned with orientation, while distal joints

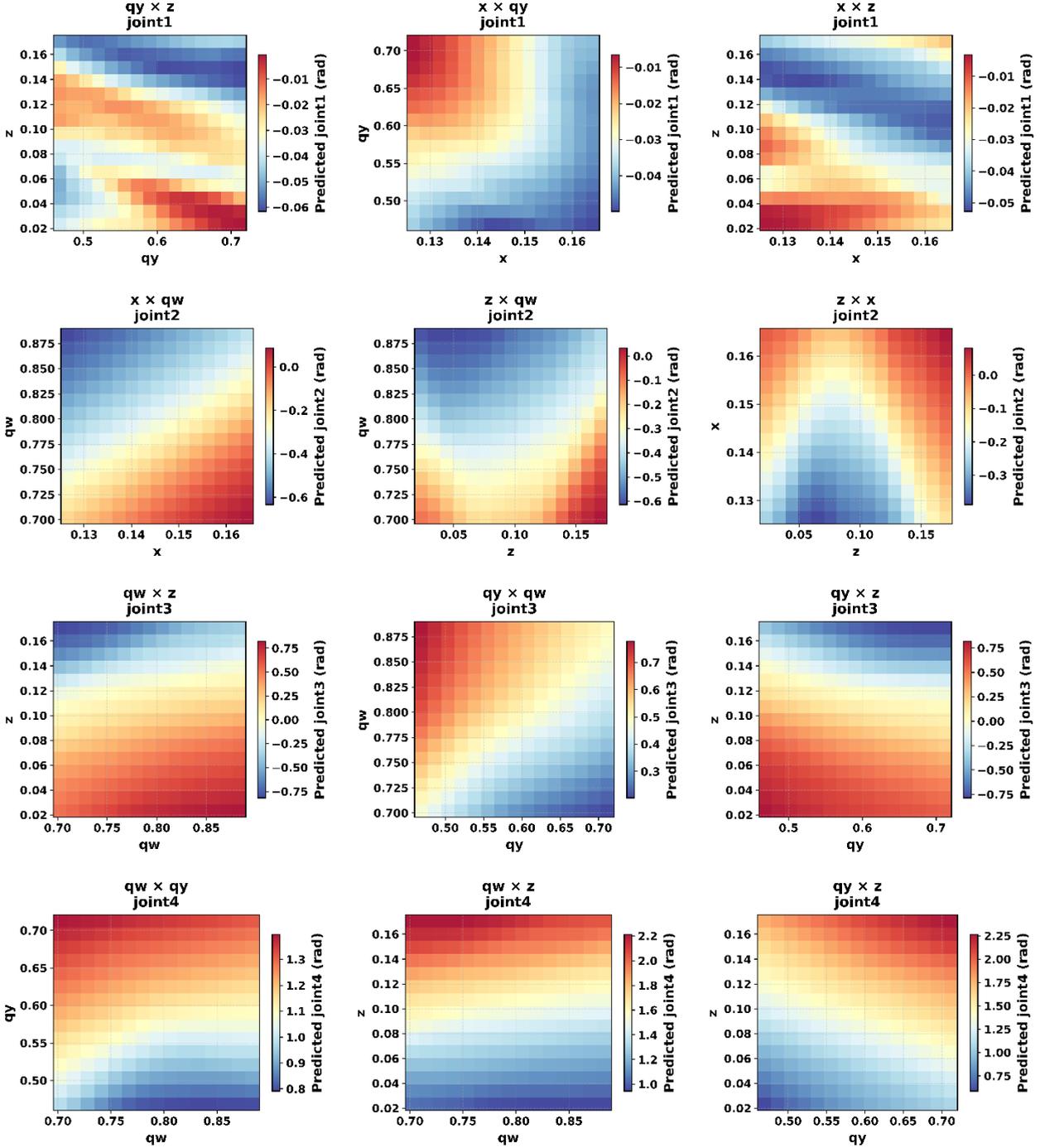

**Fig. 6.** Heat map of Original IKNet model

focus on positional adjustments, particularly along the vertical axis.

The Improved IKNet [6]'s joint-specific analysis in Fig. 9 top-right reveals a distinct pattern. Joint1 shows highest sensitivity to quaternion y (0.059), suggesting that this proximal joint prioritizes rotational adjustments. Joint2 continues this pattern with high sensitivity to quaternion w (0.105), while joints 3 and 4 maintain the predominant focus on z-position

(0.472 and 0.363 respectively). However, joint3 also shows significant sensitivity to quaternion y (0.242),

indicating a more complex role that balances both positional and rotational considerations. This multifaceted joint utilization likely contributes to the Improved IKNet [6]'s enhanced performance, enabling more sophisticated and adaptable obstacle avoidance behaviors.

For the Focused IKNet [6] in Fig 9. bottom, a similar hierarchy emerges but with different feature priorities. Joint1 shows highest sensitivity to quaternion y (0.011), but joint2 places greater emphasis on quaternion y (0.083) and z-position (0.062), suggesting a more balanced approach for this joint. Joints 3 and 4 maintain a strong focus on z-position

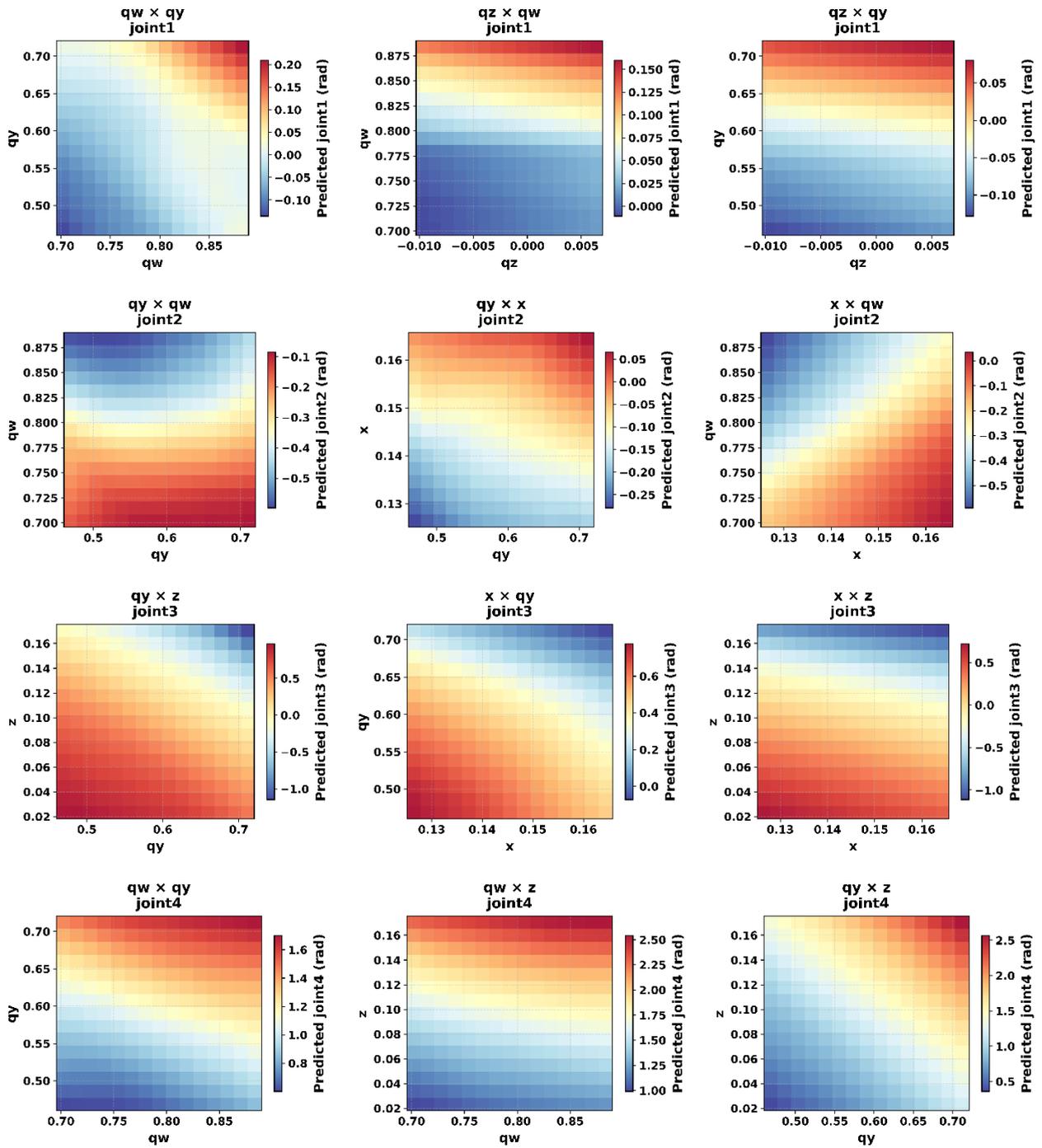

**Fig. 7.** Heat map of Improved IKNet model

(0.410 and 0.435 respectively), consistent with the other models. The quaternion y component also shows substantial importance for joint3 (0.119) and joint4 (0.114), indicating that specific rotational adjustments complement the positional strategy. This consistent emphasis on z-position for distal joints across all models highlights the fundamental.

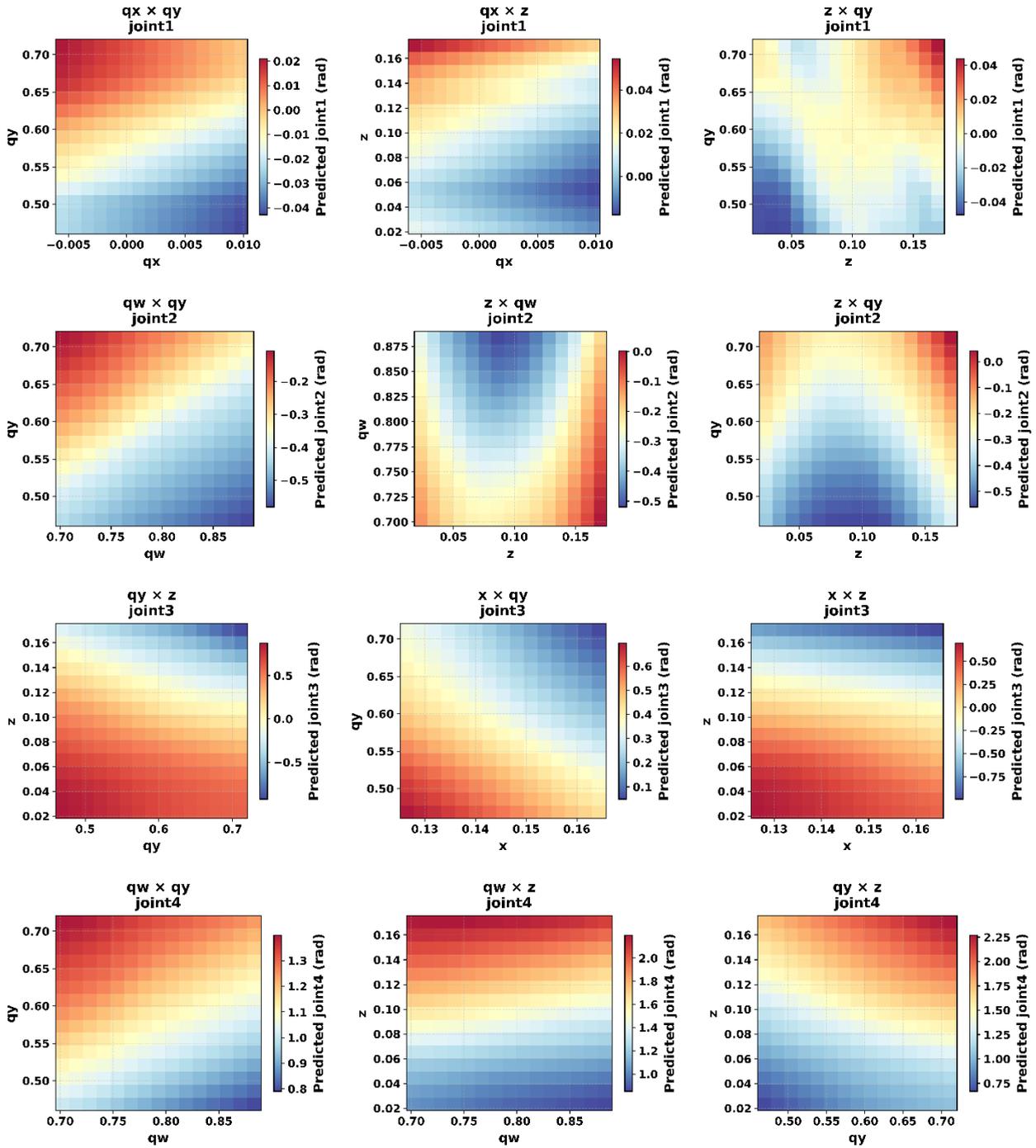

**Fig. 8.** Heat map of Focused IKNet model

### 4.3 Scenarios obstacle avoidance

The three IKNet models were evaluated across three distinct obstacle avoidance scenarios to assess their generalizability and effectiveness in different environmental contexts. These scenarios were carefully designed to present increasing levels of complexity, challenging the models with various spatial arrangements that require different navigation strategies. The visual representations of these scenarios and the models' responses provide critical insights into their practical capabilities and limitations.

#### 4.3.1 Comprehensive comparison

Fig. 10 presents a direct comparison of how the three models navigate through a representative scenario, with both 2D and 3D visualizations of their generated paths. The 2D view (left panel) clearly illustrates the efficiency difference between models, with the Improved IKNet [6] (green path) taking a more direct route compared to both the Original IKNet[5] (blue path) and Focused IKNet [6] (purple path). The Improved IKNet [6]'s trajectory shows fewer unnecessary deviations and more gradual curves, suggesting a more sophisticated path planning approach that anticipates obstacles rather than reacting to them. The Focused IKNet [6]'s path reveals sharper turns and wider berths around obstacles, indicating a more conservative approach that prioritizes clearance over path efficiency. The Original IKNet's[5] trajectory falls between these extremes, with moderate clearances but less smooth transitions than the Improved IKNet [6].

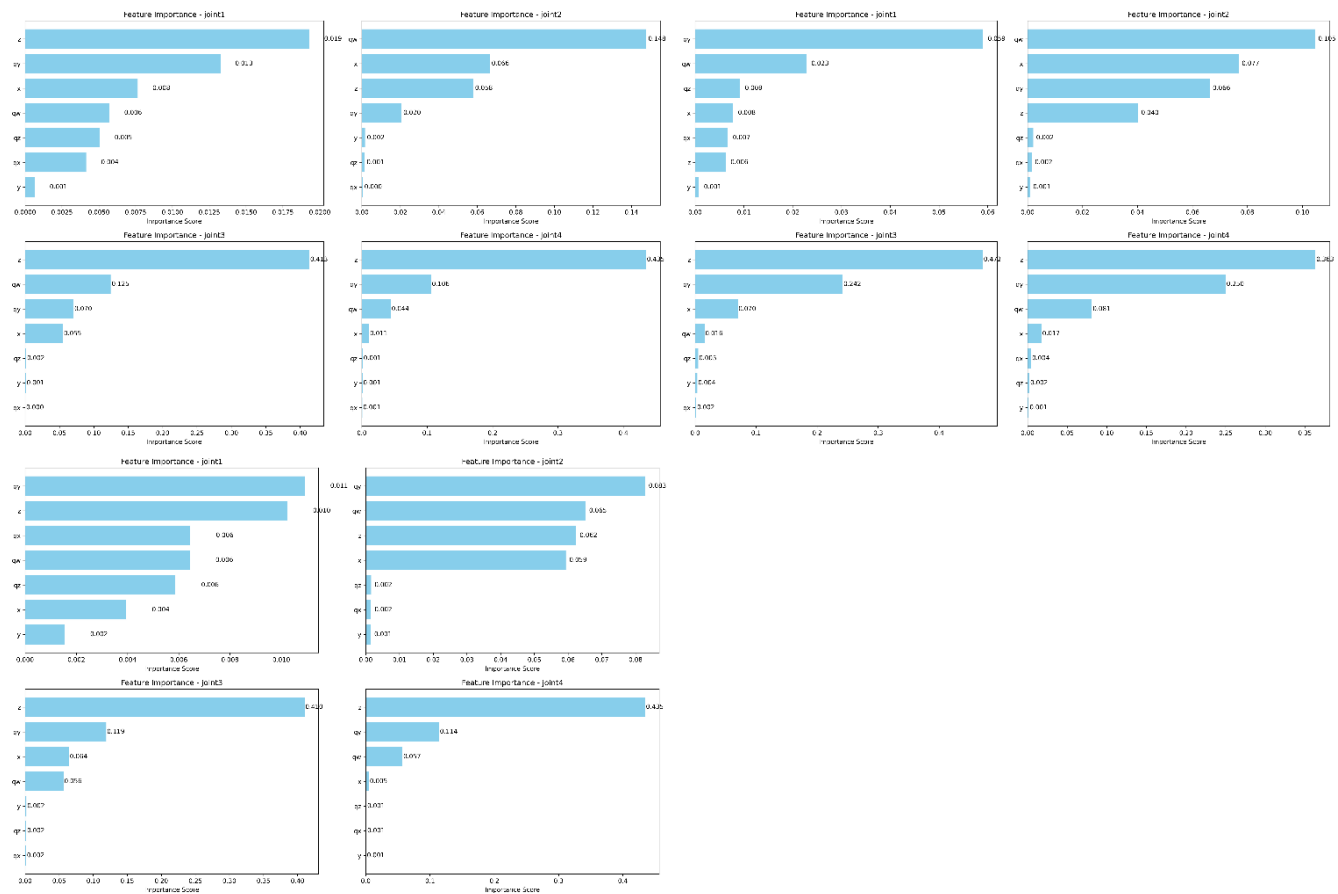

**Fig. 9.** Feature importance of three models' analysis by Interpret ML

The 3D visualization (right panel) provides additional perspective on how the models navigate the three-dimensional space, revealing that the Improved IKNet [6] maintains a more consistent altitude profile throughout its trajectory, likely contributing to its energy efficiency. The Focused IKNet [6] shows more pronounced vertical adjustments, particularly when passing near obstacles, suggesting that it actively uses the vertical dimension as part of its avoidance strategy. The Original IKNet[5] exhibits moderate altitude changes, consistent with its intermediate approach to obstacle avoidance.

Figs 11-13 provide detailed comparisons of the three models navigating through increasingly complex obstacle arrangements. Each figure presents both 2D and 3D visualizations of the models' paths, along with quantitative metrics for clearance and target error. These visualizations reveal consistent patterns in how each model approaches obstacle avoidance across different scenarios.

### 4.3.2 Scenario 1

In Fig. 11 top-left, all three models navigate a relatively simple obstacle configuration. The data boxes integrated into the visualization quantify the performance differences, with the ImprovedIKNet achieving the lowest target error (1.10 units) while maintaining an appropriate clearance (0.86 units). The FocusedIKNet demonstrates the most conservative behavior, with a clearance of 2.75 units but at the cost of a substantially higher target error (5.22 units). The Original IKNet[5] achieves intermediate values for both metrics, with a clearance of 1.67 units and target error of 2.06 units.

### 4.3.3 Scenario 2

Fig. 11 top-right presents a more challenging scenario with multiple obstacles positioned to create narrow passages. In this configuration, the Focused IKNet [6] maintains its conservative approach with generous clearances but increasingly circuitous paths. The Improved IKNet [6] continues to demonstrate superior efficiency, generating smoother trajectories that navigate the constrained space while minimizing unnecessary detours. The Original

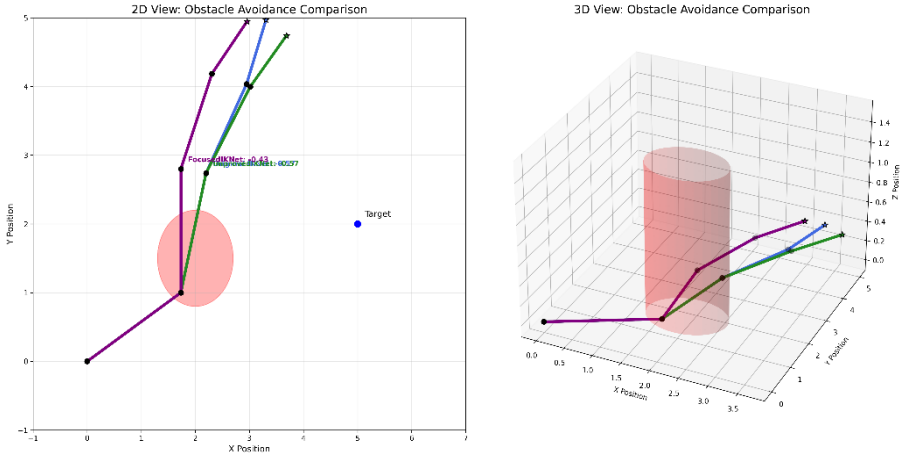

**Fig. 10.** Comprehensive comparison of obstacle avoidance

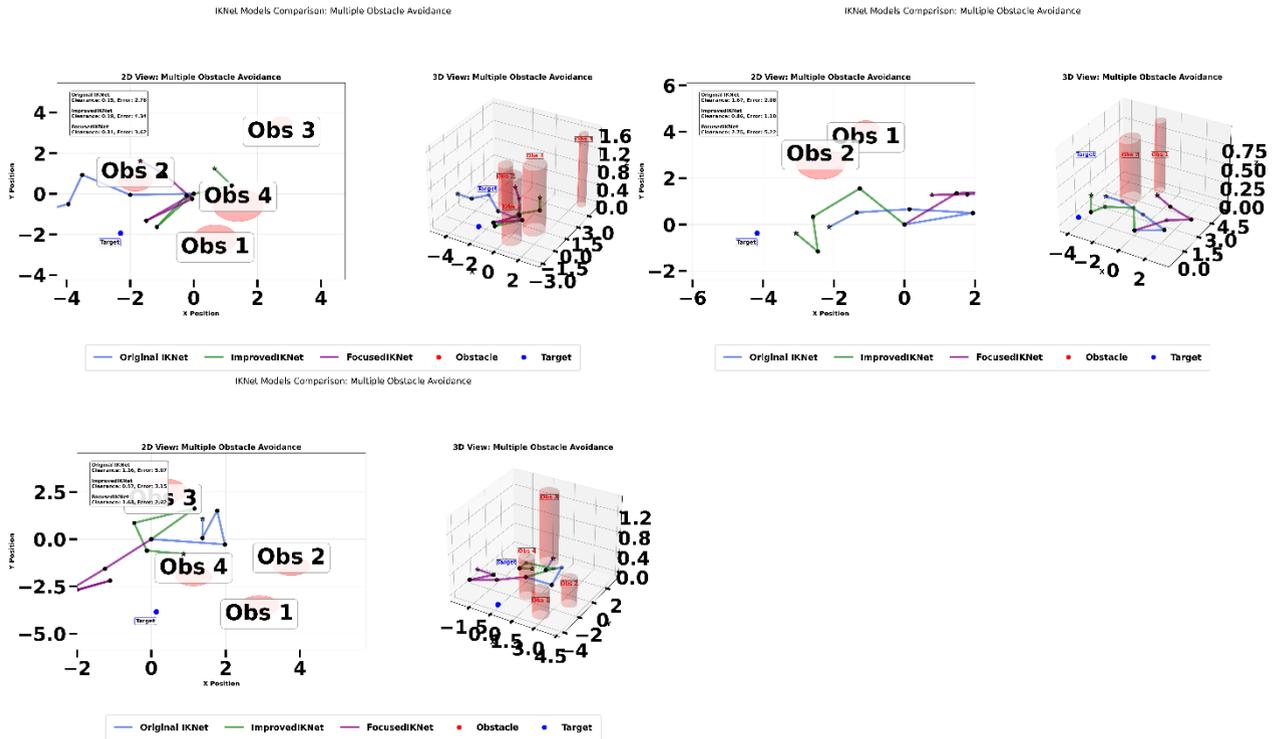

**Fig. 11.** Obstacle avoidance in each scenario

IKNet[5] shows increasing difficulty as complexity rises, with more abrupt direction changes and less optimized paths.

### 4.3.4 Scenario 3

Fig. 11 bottom showcases the most complex environment, with multiple obstacles creating a highly constrained navigation space. Here, the performance differences become even more pronounced, with the Improved IKNet [6] maintaining efficient navigation despite the increased complexity. The Focused IKNet [6]'s specialized approach shows some adaptation to the complex environment, while the Original IKNet[5] exhibits the greatest difficulty, with more erratic trajectories and higher target errors.

Across all scenarios, a consistent pattern emerges: the Improved IKNet [6] demonstrates superior path planning with smoother trajectories and better-balanced clearance-error trade-offs, the Focused IKNet [6] prioritizes safety margins at the expense of path efficiency, and the Original IKNet[5] shows moderate but less optimized performance that deteriorates with increasing environmental complexity.

### 4.4 Step by step 3 scenario obstacle avoidance

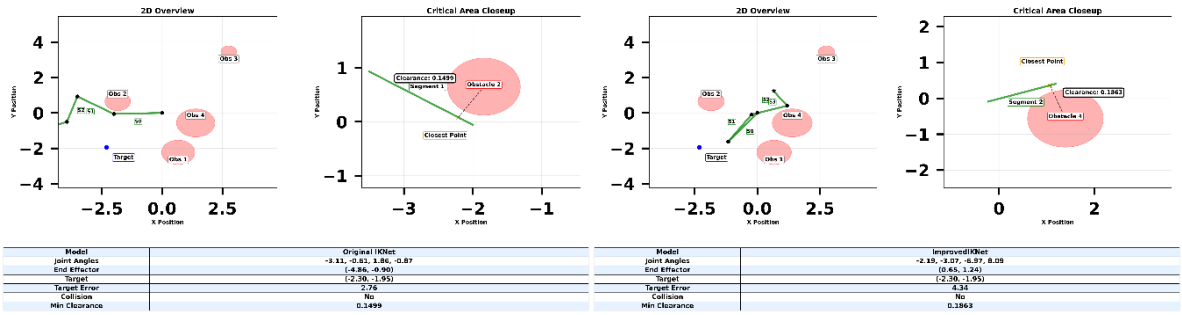

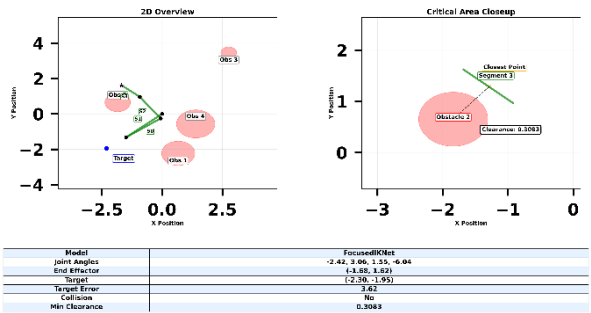

**Fig. 12.** Step by step obstacle avoidance of scenario 1

The step-by-step analysis of obstacle avoidance across different scenarios provides detailed insights into the decision-making process and trajectory generation of each IKNet[5] model. By examining the sequential states of the models as they navigate from start positions to targets, we can understand the temporal aspects of their behavior and identify key decision points that differentiate their approaches.

Figs 12-14 present detailed visualizations of how each model navigates through three increasingly complex obstacle avoidance scenarios. Each figure contains three sections representing the Focused IKNet [6], Improved IKNet [6], and Original IKNet[5], with both 2D overview and critical area closeup visualizations for each model.

### 4.4.1 Scenario 1

Fig. 12 presents a complex environment with four obstacles of different sizes positioned asymmetrically, creating a challenging navigation task. In this complex scenario, the Improved IKNet [6] achieves the most balanced performance with a minimum clearance of 0.1863 units and a target error of 4.4343 units, navigating efficiently between obstacles while maintaining a relatively direct path to the target. The Focused IKNet [6] opts for a different approach, prioritizing a wider berth around certain obstacles (clearance of 0.3083 units) at the expense of a less direct path. The Original IKNet[5] exhibits a more erratic trajectory with inconsistent clearances (0.1449 units), suggesting difficulty in handling the complexity of this scenario.

The critical area closeup panels for these scenarios highlight particularly challenging regions where the paths come close to obstacles. These detailed views reveal that the Improved IKNet [6] maintains more consistent clearances even in congested areas, while the other models show more variable behavior. The Original IKNet[5] exhibits clearances that appear minimal in some regions, potentially indicating higher collision risk under uncertain conditions or with dynamic obstacles.

### 4.4.2 Scenario 2

Fig. 13 introduces multiple obstacles with varying sizes, creating a more complex navigation environment. In this configuration, the Improved IKNet [6] demonstrates superior adaptability, adjusting its path to maintain a consistent clearance of 0.8630 units while generating a smoother trajectory that more directly approaches the target. The path exhibits fewer sharp turns and more gradual transitions, indicating more sophisticated motion planning that anticipates the entire trajectory rather than responding reactively to each obstacle. The Focused IKNet [6] takes a relatively conservative approach, generating a path that maintains substantial distance from both obstacles but results in a less direct route to the target. The Original IKNet[5] shows an intermediate approach, achieving a clearance like the Focused IKNet [6] but with more abrupt path transitions, particularly evident in the sharp angle formed near the first obstacle.

### 4.4.3 Scenario 3

In Fig. 14, a single obstacle is positioned between the robot and the target, creating a basic avoidance challenge. The 2D overview panels show the complete spatial arrangement, with pink ellipses representing obstacles, blue dots indicating targets, and green lines showing the models' generated paths. The Focused IKNet [6] generates a path that makes a relatively sharp turn around the obstacle, maintaining a clearance of approximately 1.5604 units as indicated in the data table beneath the visualization. The Improved IKNet [6] generates a more efficient path, with smoother transitions between segments and a clearance of approximately 0.5402 units, suggesting better optimization between safety and efficiency. The Original IKNet[5] produces a path with moderate efficiency, maintaining a clearance of approximately 0.9932 units but with somewhat less smooth transitions.

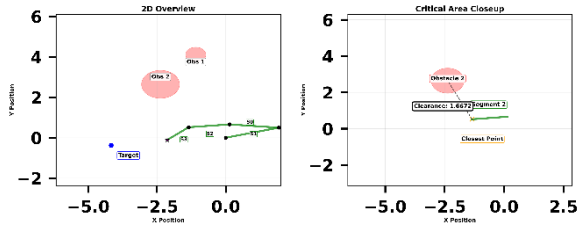
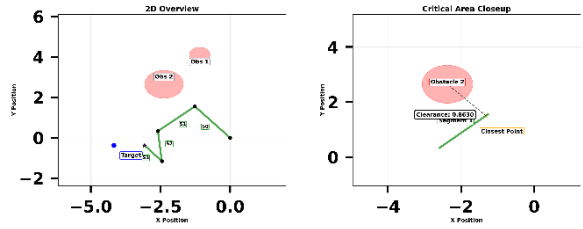
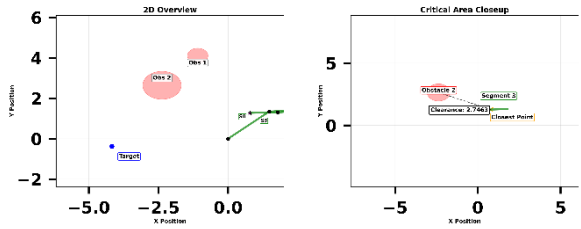

**Fig. 13.** Step by step obstacle avoidance of scenario 2

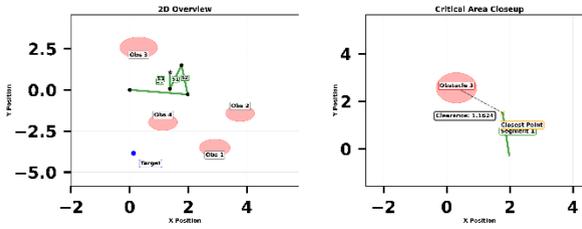
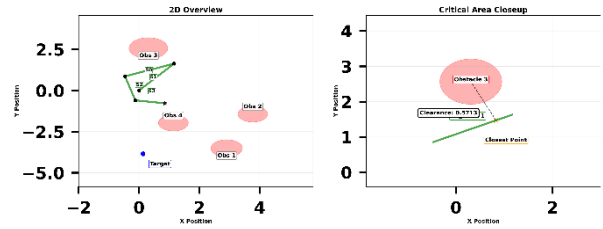
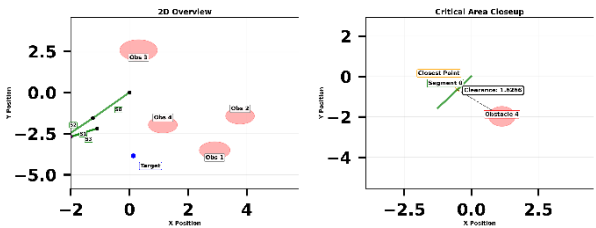

**Fig. 14.** Step by step obstacle avoidance of scenario 3

The critical area closeup panels provide magnified views of the regions where the paths come closest to obstacles, enabling detailed examination of the models' clearance behaviors. These closeups reveal that the Improved IKNet [6] maintains a consistent minimum clearance with a better approach angle, suggesting improved spatial awareness and more sophisticated path planning. The data tables below each visualization provide quantitative metrics, confirming that the Improved IKNet [6] achieves the lowest target error (2.8651) while maintaining appropriate clearance, indicating superior overall performance in this basic scenario.

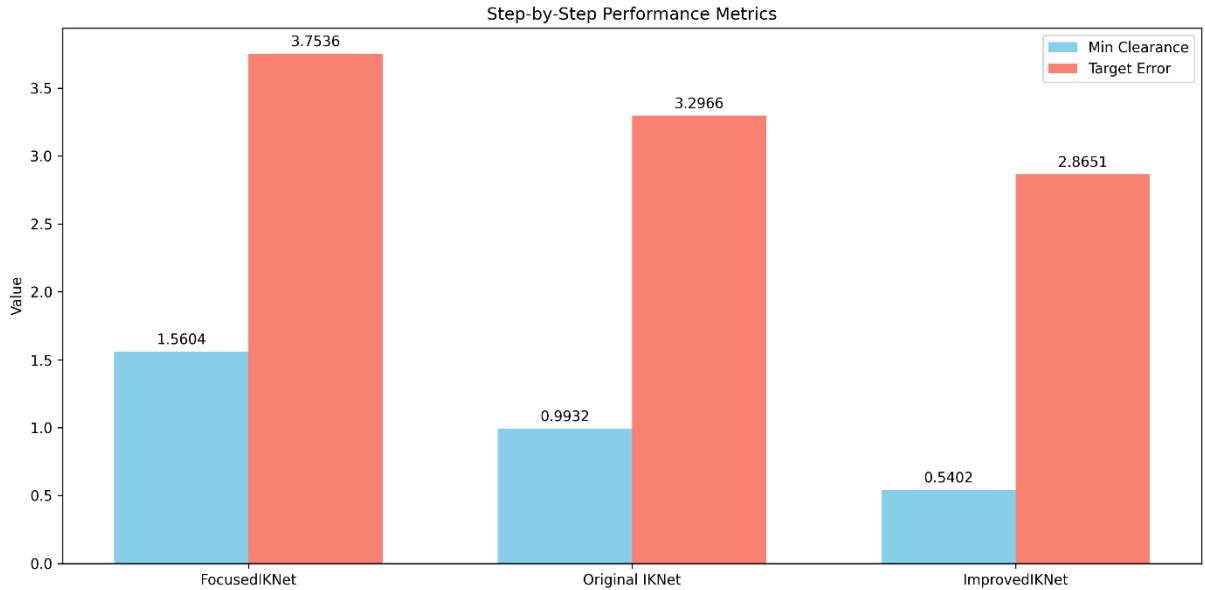

**Fig. 15.** Step by step obstacle avoidance summary

### 4.4.4 Summary of step-by-step obstacle avoidance

Fig. 15 provides a visual comparison of the key performance metrics across the three models. This bar chart clearly illustrates the trade-offs between minimum clearance (blue bars) and target error (red bars) for each model. The Improved IKNet [6] demonstrates the optimal balance, with the lowest values for both metrics, indicating efficient path planning that maintains appropriate safety margins while minimizing unnecessary deviations. The Focused IKNet [6] shows the highest target error and clearance values, reflecting its conservative approach that prioritizes obstacle avoidance over path efficiency. The Original IKNet[5] displays intermediate values, suggesting a moderate but less optimized approach to the obstacle avoidance task.

The step-by-step analysis reveals important temporal aspects of the models' behavior. The Improved IKNet [6] consistently demonstrates anticipatory path planning, initiating gradual course adjustments well before reaching obstacle proximities. This forward-looking approach results in smoother trajectories with fewer abrupt direction changes, suggesting more sophisticated spatial reasoning that considers the entire environment rather than responding reactively to immediate obstacles. The Focused IKNet [6] exhibits more conservative behavior, making earlier and more pronounced deviations to maintain generous clearances from obstacles. The Original IKNet[5] shows less consistent behavior, with some anticipatory adjustments but also more reactive movements when approaching obstacles.

This detailed analysis across the three scenarios confirms the ImprovedIKNet's superior performance in obstacle avoidance tasks, demonstrating its ability to generate efficient, smooth trajectories that balance safety considerations with path optimization. The consistent pattern of performance across scenarios with increasing complexity suggests better generalizability and robustness, qualities that are essential for real-world robotic applications where environmental conditions may vary unpredictably.

## 5 Summary and conclusion
### 5.1 Summary

The comprehensive analysis of the three IKNet variants—Original IKNet[5], Improved IKNet [6], and Focused IKNet [6]—reveals significant differences in their obstacle avoidance strategies, feature utilization patterns, and overall performance across various scenarios. These differences highlight the impact of model architecture and training approach on robotic navigation capabilities and provide valuable insights for the development of more effective inverse kinematics solutions for obstacle avoidance tasks.

Fig. 16 provides a consolidated summary of the XAI (explainable AI) analysis, presenting key performance metrics and feature importance rankings for all three models in a comparative format. This tabular representation includes top features based on SHAP analysis, top features based on custom analysis methodologies, obstacle clearance, target

## XAI Analysis Summary

| Model | Top Features (SHAP) | Top Features (Custom) | Obstacle Clearance | Target Error | Collisions |
|---|---|---|---|---|---|
| FocusedIKNet | qz, qw, y | qy, z, qx | 1.5604 | 3.7536 | No |
| ImprovedIKNet | qz, qw, qy | qy, qw, qz | 0.5402 | 2.8651 | No |
| Original IKNet | qz, qw, qx | z, qy, x | 0.9932 | 3.2966 | No |

**Fig. 16.** XAI Summary

error, and collision occurrences for each model. The Improved IKNet [6] demonstrates superior performance with the lowest average target error (2.8651 units) and the significant advancement in navigation efficiency, potentially translating to reduced energy consumption and faster task completion in real-world robotic applications. The Original IKNet[5] shows moderate performance across metrics, with a target error of 3.2966 units and clearance of 0.9932 units, suggesting a functional but less optimized approach to obstacle avoidance. The Focused IKNet [6], despite its specialized approach, records higher target errors (3.7536 units) while maintaining larger clearances (1.5604 units), indicating a strategy that prioritizes safety margins at the expense of path efficiency.

The absence of collisions across all models, as indicated in Fig. 16, confirms that all three approaches successfully achieve the fundamental safety requirement of obstacle avoidance. However, the variations in clearance and target error metrics reveal important differences in how efficiently they accomplish this goal, with direct implications for energy efficiency, task completion time, and overall system performance in practical applications.

The SHAP analysis provides valuable insights into the decision-making processes of these models, revealing fundamental differences in how they prioritize and utilize different kinematic features. The Improved IKNet [6] exhibits more balanced feature utilization, integrating both positional and quaternion-based information effectively. This balanced approach contributes to its superior performance, enabling more sophisticated spatial reasoning that considers multiple aspects of the robot's configuration simultaneously. The more uniform sensitivity across feature spaces observed in the heat maps further confirms this balanced approach, suggesting that the Improved IKNet [6] responds more consistently to variations in different input features.

The Original IKNet[5] shows higher dependency on specific quaternion components, particularly for joints 3 and 4, indicating a more specialized approach that may perform well in certain configurations but less optimally in others. The concentrated regions of sensitivity in its heat maps smallest minimum clearance (0.5402 units), indicating efficient path planning that balances target accuracy with obstacle avoidance. This optimal balance represents a suggest that the model may be highly responsive to input patterns but less adaptable to variations outside these optimized regions. This characteristic may explain its moderate performance across diverse scenarios, as it may lack the flexibility to adapt optimally to varying environmental configurations.

The Focused IKNet [6] demonstrates the most specialized feature importance distribution, with particularly high weights assigned to specific positional and quaternion components. This highly targeted approach may enable exceptional performance in scenarios that align well with its specialized strategy but may limit its generalizability across diverse environments. The concentrated sensitivity patterns observed in its heat maps reinforce this characterization, suggesting a model that has developed highly specific responses to spatial arrangements.

### 5.2 Conclusion

In conclusion, this analysis demonstrates that the Improved IKNet [6] represents a significant advancement over the Original IKNet[5], particularly in complex obstacle avoidance scenarios. The model's balanced feature utilization, smooth trajectory generation, and optimal clearance-error trade-off position as the preferred choice for robotic applications requiring efficient navigation in cluttered environments. The comprehensive evaluation across multiple scenarios with varying complexity confirms the robustness of this improvement, suggesting that the architectural and training enhancements implemented in the Improved IKNet [6] have successfully addressed limitations in the original model. In summary, the best choice of model doing obstacle avoidance will be Improved IKNet[6].